\documentclass[letterpaper,twocolumn,10pt]{article}
\usepackage{usenix2019_v3}
\usepackage[available,functional,reproduced]{usenixbadges}

\usepackage{epsfig,endnotes}
\usepackage[T1]{fontenc}
\usepackage[scaled=0.75]{beramono}
\usepackage{amsmath,amsthm}
\usepackage{mathptmx}
\usepackage[ruled, noend, vlined]{algorithm2e}
\usepackage{listings}
\usepackage{xcolor}
\usepackage{colortbl}
\usepackage{balance}
\usepackage{graphicx}
\usepackage[font={small}]{subcaption}
\usepackage{wrapfig}
\usepackage[font={small}]{caption}
\usepackage{booktabs}
\usepackage[export]{adjustbox}
\usepackage{datetime}
\usepackage{enumitem}
\usepackage{multirow}
\usepackage[binary-units=true]{siunitx}
\usepackage{pifont}
\usepackage{tikz}
\usetikzlibrary{tikzmark,calc}
\usepackage{stackengine}
\usepackage[nameinlink]{cleveref}
\usepackage[title]{appendix}
\usepackage[compact]{titlesec}
\usepackage[english]{babel}
\usepackage{soul}

\graphicspath{{./}{figures/}}

\DeclareMathAlphabet{\mathcal}{OMS}{cmsy}{m}{n}

\newcommand{\sys}{\textsc{TrainCheck}\xspace}
\newcommand{\tool}{\textsc{TrainCheck}\xspace}
\newcommand{\instrumentor}{Instrumentor\xspace}
\newcommand{\inferengine}{Infer Engine\xspace}
\newcommand{\checker}{Verifier\xspace}

\newcommand{\numofbugs}{88\xspace}
\newcommand{\numofnewbugs}{6\xspace}

\newcommand{\numofnewbugsfixed}{3\xspace}
\newcommand{\numofevalsetbugs}{20\xspace}
\newcommand{\numofevalsetbugsdetected}{18\xspace}
\newcommand{\numofevalfpprogs}{63\xspace}

\newcommand{\inlinecode}[1]{\texttt{#1}}

\DeclareSIUnit{\operation}{op}
\definecolor{ngray}{RGB}{102,102,102}

\newcolumntype{R}[2]{%
  >{\adjustbox{angle=#1,lap=\width-(#2)}\bgroup}%
  l%
  <{\egroup}%
}

\definecolor{bg}{rgb}{0.95,0.95,0.95}

\crefname{section}{\S}{\S\S}
\crefname{subsection}{\S}{\S\S}
\crefformat{section}{\S#2#1#3}
\crefformat{subsection}{\S#2#1#3}

\newcounter{magicrownumbers}

\addto\extrasenglish{%
}

\begin{document}
\hypersetup{
  colorlinks=true,
  linkcolor=purple,
  citecolor=blue,
  urlcolor=blue,
}

\date{}
\title{\LARGE Training with Confidence: Catching Silent Errors in Deep Learning Training with Automated Proactive Checks}
\author{
{\rm Yuxuan Jiang}\and
{\rm Ziming Zhou}\and
{\rm Boyu Xu}\and
{\rm Beijie Liu}\and
{\rm Runhui Xu}\and
{\rm Peng Huang}\and \\
University of Michigan \\
 \addlinespace[0.5cm]
 {
\rm
Technical Report\thanks{A shorter version of this paper is to appear in the Proceedings of the
19th USENIX Symposium on Operating Systems Design and Implementation (OSDI '25), July, 2025}
 }
}


\maketitle

\begin{abstract}
Training deep learning (DL) models is a complex process, making it prone to
silent errors that are challenging to detect and diagnose. This paper
presents \sys, a framework that takes a proactive checking approach to
address silent training errors. \sys automatically infers invariants tailored
for DL training. It uses these invariants to proactively detect silent errors
during the training process while providing debugging help. To evaluate \sys,
we reproduce 20 \emph{real-world} silent training errors with diverse root
causes. \sys successfully detects 18 errors within a single training iteration.
It also uncovers 6 unknown bugs in popular training libraries that lead to
silent errors.
\end{abstract}

\section{Introduction}
\label{sec:intro}
Training deep learning (DL) models has become integral for many application
domains~\cite{lecun_deep_2015,esteva_guide_2019,bommasani2022opportunitiesrisksfoundationmodels}.
DL training, however, is a complex process involving multiple steps and layers
of components such as user code, compiler, training framework, optimization
libraries, drivers, and distributed systems. Moreover, these components undergo
frequent updates~\cite{Dilhara2021MLSoftware,Chen2023DLFrameworkBugs} due to the
rapid pace of DL research. Consequently, training jobs are prone to errors from
various sources~\cite{DLFailure2020ICSE}.

To make matters worse, while some errors cause immediate job failures and are
relatively easy to identify (\emph{e.g.}, GPU out of memory or illegal argument
exceptions), many others are \emph{silent or latent}. Such errors do not cause
obvious training disruptions but eventually produce suboptimal/incorrect models
or cause noticeable failures much later.

\begin{figure}[t]
  \centering
  \includegraphics[width=3.25in]{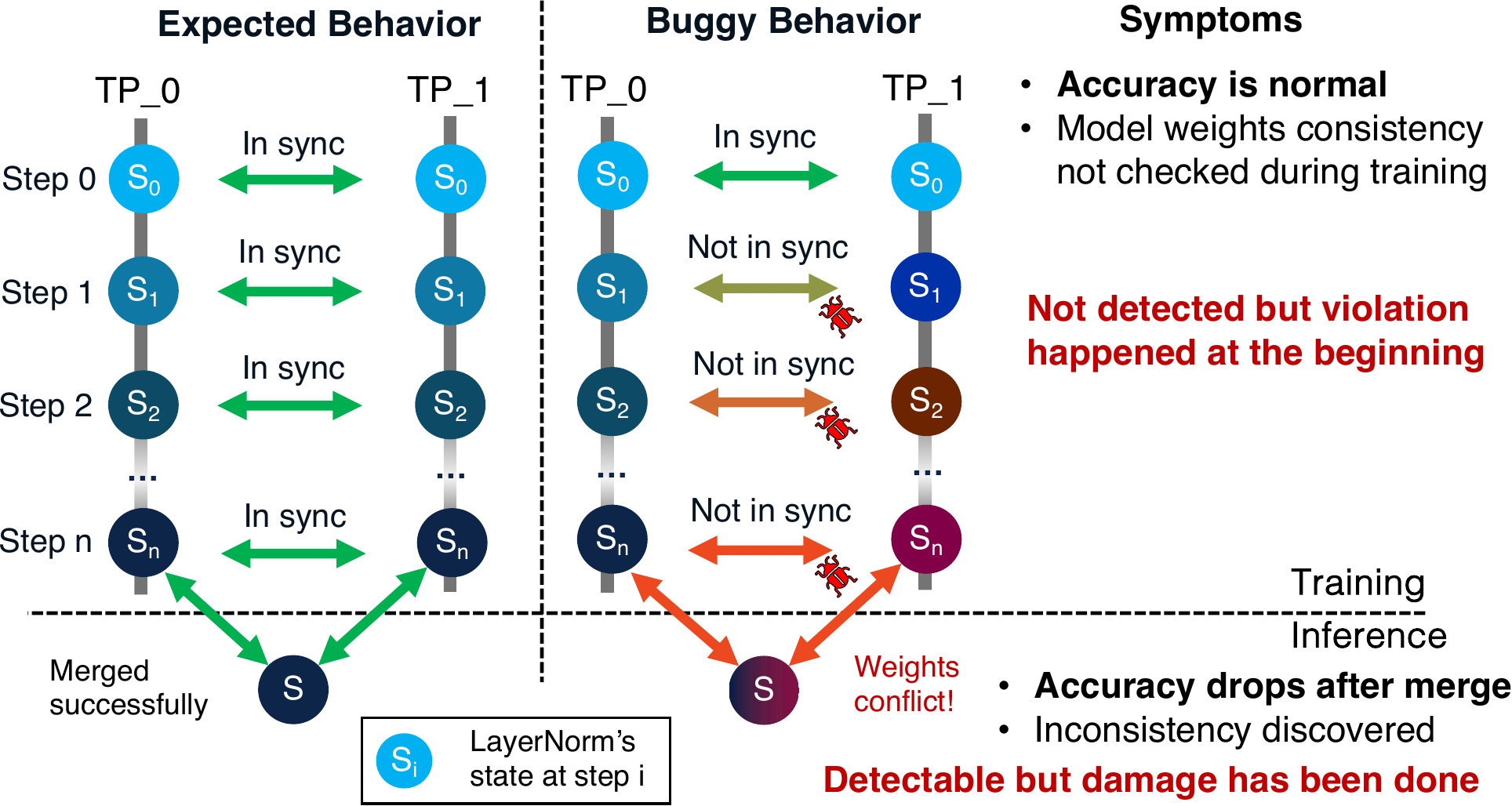}
  \caption{Silent error in BLOOM-176B training.}
  \label{fig:ds_1801}
  \vspace{-0.1in}
\end{figure}

\autoref{fig:ds_1801} shows a \emph{real-world} silent training error in
HuggingFace's training of BLOOM-176B~\cite{huggingface2022bloom}. DeepSpeed's
\texttt{BF16Optimizer} was used in this training task, which had a logic bug in
gradient clipping.  This bug did not trigger any exception but caused parts of
the model to diverge silently across GPUs, a problem that went undetected for 10
days.

Detecting such silent errors is inherently difficult due to the lack of clear
signals.  When they are detected, substantial training resources have been
wasted. The problem is further compounded by the trend of training large models
using massive resources. Moreover, during diagnosis, developers are often left
in the dark and have to take a trial-and-error approach~\cite{DLStudy2019ISSRE},
such as tweaking hyperparameters and rerunning the task while the real root
causes are elsewhere, which is tedious and time-consuming. Indeed,
developers have expressed frustration with diagnosing these silent
errors~\cite{wu2017mistakes, wu2020silentbugs, reddit_pytorch_numpy_bug,
tanelp_numpy_bug, stackoverflow_dataloader_sampling,rishy_dropout_theano}.

Current practices rely on high-level model evaluation signals such as loss,
accuracy, and gradient norms~\cite{wandb,TensorFlow2016OSDI}. While these
signals provide useful information about overall training progress, they are not
designed for error detection. They are often noisy and evaluated only
periodically, leading to missed or delayed detection. Even when anomalies are
observed, they provide little diagnostic value for identifying the root cause.
The aforementioned BLOOM-176B training error, for example, did not manifest as
abnormal loss or accuracy, and it took developers 12 days to diagnose and
resolve the problem~\cite{bigscience2022hanging,huggingface2022bloom}. A few
static solutions exist to detect certain errors with specific root causes, such
as tensor shape mismatches~\cite{PyTea2022ICSE}, but they fail to cover many
other types of silent training errors.

Our insight is that a fundamental reason behind the challenges is the lack of
\emph{proactive checks} to continuously validate a training task. We call such
checks, \emph{training invariants}, which are rules that should hold throughout
the training. Violation of a training invariant indicates potential errors. The
concept of invariants is not new and has been widely studied in traditional
software. However, existing tools such as Daikon~\cite{Daikon1999ICSE} focus on
low-level variable relationships (e.g., \inlinecode{var1 > var2}) and fail to
capture the high-level semantics of silent errors in DL training, making them
ill-suited for this domain.

Based on this insight, we design an end-to-end framework, \sys, that enhances a
DL training job with training invariants to proactively detect a wide variety of
silent errors. Compared to generic high-level signals, the training invariants
\sys introduces can more accurately and quickly detect anomalies. Upon
detection, \sys provides debugging information indicated through the violated
invariant and relevant traces. Compared to static approaches, \sys can provide
stronger assurance. Its checks capture misbehavior caused by various root
causes. It operates continuously with a running training task, enabling the
detection of real errors in deployment, which may only occur under specific
datasets, environments, or scales.

In designing \sys, we aim to explore two key research questions. First,
\emph{what kind of training invariants are effective} to address silent training
errors? We initially hypothesized that, due to the stochastic nature of DL
training, the invariants would need to encode probabilistic predicates, which
can be complex and unstable. Our further investigation leads to the insight
that, by properly choosing the level of behavior to observe, a training
invariant can be made simpler and more precise. Non-determinism is an artifact
of checking at too high of a level. The invariants should operate at a level
below model evaluation signals, but not as low as traditional software
invariants. In the BLOOM-176B example, an effective training invariant is
roughly: \ul{the weights of certain layers should stay consistent across tensor
parallelism (TP) ranks}. We thus focus on such rule-level training invariants.

Second, \emph{how can we automatically obtain and check training invariants}?
The above invariant description is informal. The actual invariant must be
concrete and detailed enough to be directly \emph{checkable} for a specific
training task. It is impractical to ask developers to manually write and
maintain such invariants. Deep learning frameworks and training practices evolve
rapidly, introducing new APIs, model architectures, and algorithms that make
handwritten invariants difficult to catch up. Moreover, invariants must match
precisely with the actual implementation, not just high-level intent. 

Automated inference is needed, but this is challenging.  For example, to infer
the above invariant, we need to collect information about each worker's role
(\emph{e.g.}, TP rank), tensor properties (replicated or partitioned), and
observed values.  The conditions under which this invariant applies are subtle:
it only holds in distributed training when using tensor parallelism and
LayerNorm, and only for weights \emph{replicated} across TP ranks.  This makes
identifying the correct \emph{precondition} critical---for example, the tensor
must have \texttt{tensor\_model\_parallel=False}.  These kinds of invariants go
beyond what traditional invariant inference tools like
Daikon~\cite{Daikon1999ICSE} are designed to handle.

\sys is designed to address these unique challenges and automatically infer
concrete, checkable training invariants for assorted DL training tasks written
on top of popular frameworks such as PyTorch~\cite{Ansel_PyTorch_2_Faster_2024}
and DeepSpeed~\cite{deepspeed}.  \sys first instruments a given DL training
program to collect traces. To achieve high usability and low overhead, we take a
monkey-patching approach to dynamically inject the instrumentation code for
framework API invocations and a proxy-based approach to intercept state updates.

To infer invariants from the collected traces, \sys defines a set of generic
relation templates.  It uses an efficient algorithm that generates hypotheses
based on a relation template and validates the hypotheses in the traces to
generate invariants. \sys further designs an algorithm to deduce the
precondition, if any, for each invariant.

To use the inferred invariants for detecting silent training errors in a
specific training pipeline, \sys selectively instruments the pipeline for only
information relevant to the invariants.  A verifier continuously validates the
traces from the instrumented training task to check violations.

Unlike traditional invariant inference tools that operate within a single
program, a unique feature of \sys is its ability to generate invariants
\emph{transferable} across different training programs and even different
libraries. This makes \sys broadly applicable and adaptable. Consequently, we
can leverage high-quality DL training pipelines, such as those found in
tutorials and example repositories, \emph{e.g.}, PyTorch
examples~\cite{pytorch-examples}, to infer the invariants and apply them to
other training programs. This helps with both improving invariant accuracy and
aggregating effective invariants.

For evaluation, we collect and reproduce 20 \emph{real-world} silent training
errors with diverse root causes. \sys detects \numofevalsetbugsdetected cases
within a single training iteration while providing debugging hints. \sys
additionally uncovers \numofnewbugs previously unknown errors, all of which have
been confirmed with \numofnewbugsfixed being fixed.  We also share our
experiences applying \sys to various DL training scenarios and highlight
challenges and opportunities in this domain. \sys is open sourced at
\url{https://github.com/OrderLab/TrainCheck}.

The contributions of this paper are as follows:
\begin{itemize}[noitemsep, topsep=0pt, partopsep=0pt, leftmargin=*]
  \item We investigate the notorious yet under-explored problem of silent errors
    in DL training and conduct an empirical study to shed light on their
    characteristics.
  \item We propose an approach that uses training invariants to proactively
    validate DL training and catch silent errors.
  \item We design and implement \sys, which, to the best of our knowledge, is
    the first framework that automatically infers and checks invariants tailored
    for DL training tasks.
  \item We evaluate \sys in detecting real-world silent training errors and
    exposing unknown errors.
\end{itemize}

\section{Silent Errors in DL Training}
\label{sec:ml-silent-issues}
To gain a deeper understanding of silent training errors, we conduct an
empirical study on \emph{real-world} errors that DL practitioners encountered.
We aim to provide insights regarding their root causes, impact, and current
detection methods.

\paragraph{Methodology}
To ensure the study is representative, we inspect diverse sources: (1) GitHub
issues from popular libraries such as PyTorch~\cite{pytorch_github_issues} and
DeepSpeed~\cite{deepspeed_github_issues}, (2) discussion forums such as
StackOverflow~\cite{stackoverflow} and the PyTorch
forums~\cite{pytorch_discussion_forum}, and (3) papers and blog posts from DL
practitioners~\cite{bigscience2022hanging,zhang2022optopenpretrainedtransformer}.
For GitHub, we search closed issues labeled or containing the keywords
``correctness'' or ``silent''. For discussion forums, we search for posts with
``silent'' or ``bug'' in the title or description. We also examine DL training
papers and blog posts from large organizations such as
HuggingFace~\cite{huggingface2022bloom},
Bloomberg~\cite{wu2023bloomberggptlargelanguagemodel}, and
Meta~\cite{dubey2024llama3herdmodels,zhang2022optopenpretrainedtransformer},
particularly those discussing widely deployed LLMs like
Llama3~\cite{dubey2024llama3herdmodels},
OPT~\cite{zhang2022optopenpretrainedtransformer}, and
BLOOM~\cite{huggingface2022bloom}. These sources often describe errors arising
in industrial-scale training pipelines. Many of the candidate reports from
GitHub and user forums are incomplete or poorly understood, as silent errors
are often difficult to diagnose. We carefully inspect each case and curate a set
of high-quality instances that are well-documented, diagnosed with clear root
causes, likely reproducible, and impactful.

In total, we collect~\numofbugs silent errors with known root causes.  Among
the sources, 70 are GitHub repository issues, 16 from discussion forums
such as StackOverflow and the official PyTorch Forums, and 2 from industrial
reports~\cite{huggingface2022bloom,zhang2022optopenpretrainedtransformer}.

\subsection{Analyses and Observations}

\begin{figure}[t]
  \begin{subfigure}[b]{0.35\linewidth}
    \centering
    \includegraphics[width=\linewidth]{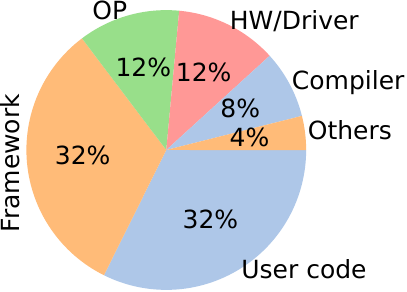}
    \caption{Location.}
    \label{fig:finding:location}
  \end{subfigure}%
  ~
  \begin{subfigure}[b]{0.63\linewidth}
    \centering
    \includegraphics[width=\linewidth]{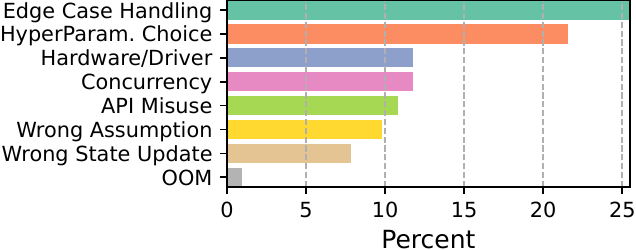}
    \caption{Type.}
    \label{fig:finding:cause}
  \end{subfigure}
  \caption{Root cause locations and types of the studied errors.}
\end{figure}

\paragraph*{Diverse Root Causes} The studied errors are caused by defects in
a wide range of components involved in DL training, including user code,
framework, compiler, mathematical operators for different hardware or from optimization
libraries, and underlying system components such as driver or hardware.

\autoref{fig:finding:location} shows the distribution of the locations of the
root causes. The majority of the errors are caused by user code (32\%) and
framework (32\%), followed by mathematical operations (12\%), hardware (12\%),
compiler (8\%), and other factors (4\%).  User code may contain missing or
incorrect API calls or poorly chosen hyperparameters. Frameworks may have bugs
in their Python components, often involving high-level algorithm logic.
Mathematical operations such as matrix multiplication can produce inaccurate
results. Hardware failures are usually caused by driver or device faults that
lead to communication errors or memory corruption. Compiler failures occur when
Torch Dynamo, PyTorch's JIT compiler, is unable to compile the pipeline
correctly. 

Figure~\ref{fig:finding:cause} shows the root cause types for the
studied cases. For user-code-induced silent errors, we identify two
primary categories: (1) \emph{Incorrect Implementation} and (2)
\emph{Inappropriate Design Choices}. The first includes incorrect, missing, or
inconsistent API calls or parameter updates. For example, one error is caused
by a missing \inlinecode{zero\_grad()} inside the training loop, resulting in
noisy gradients; another stems from initializing the optimizer before model
transformations, leaving the optimizer with incorrect parameters to update. The
second category involves flawed design choices, such as selecting an unsuitable
loss function, setting an overly aggressive dropout rate, or using a problematic
data processing pipeline. While individual components may appear correct, their
interaction often leads to numerical instability or divergence. For instance,
BLOOM developers encountered repeated loss explosions when training with the
float16 data type, which was mitigated by switching to bfloat16 and applying
gradient clipping.

\begin{table}[t]
  \small
  \centering
  \begin{tabular}{@{}lcccc@{}}
    \toprule
    \textbf{Iter} & \textbf{Type} & \textbf{Loss Diff} & \textbf{PPL Diff} & \textbf{Diff (Loss/PPL)} \\
    \midrule
    2000 & Valid & +1.14\% & +1.43\% & +0.014 / +0.050 \\ 
    2000 & Test  & +2.74\% & +3.30\% & +0.032 / +0.107 \\ 
    4000 & Valid & +3.05\% & +3.36\% & +0.033 / +0.099 \\ 
    4000 & Test  & +4.67\% & +4.79\% & +0.047 / +0.131 \\ 
    \bottomrule
  \end{tabular}
  \caption{Reproducing DeepSpeed-1801 (the root cause of BLOOM-176B silent training error)
  in a small transformer-based language model. %
  Due to resource constraints, we trained for only 2000 and 4000
  iterations.  However, the impact of the error---observed as differences in
  loss and perplexity due to weight merging---is already noticeable and
  increases with the number of training iterations.}
  \label{tab:bug-impact-ds-1801}
\end{table}

\paragraph{Severe Consequences} The impact of silent errors can be enormous. DL
training tasks are often run on many GPUs for a long time. This is the case for
the motivating example in~\autoref{fig:ds_1801}, which is a large-scale
training task that involves 384 A100 GPUs for 3.5 months. Discovering errors late
in the training result in significant waste of expensive resources.

The majority of the studied errors result in either suboptimal or incorrect
models. In the former case, users may choose to accept the degraded outcome to
avoid paying the expensive retraining cost, but must live with lower model
performance. In the latter case, rerunning the job leads to a significant waste
of costly training resources. Many errors also introduce severe performance
impacts. For example, in PyTorch-Forum-84911, the data processing code in the
user program mistakenly resizes the input images to 1024\,$\times$\,1024 instead
of the expected 224\,$\times$\,224, significantly increasing per-iteration
training time. Moreover, due to their silent nature, the effects of these errors
often persist and accumulate over time.

\paragraph{Challenging Detection and Diagnosis} Unlike other DL training
failures that trigger explicit exceptions (\emph{e.g.}, Illegal Argument) or
terminate jobs, silent training errors are challenging to detect. In many of our
studied errors, although the root cause is triggered early on, the error remains
undetected for a long time. Developers currently mainly rely on high-level
model evaluation metrics such as loss or accuracy. However, silent errors often
do not cause immediate anomalies in such evaluation metrics. For example, in
PyTorch-Forum-84911, the loss and accuracy metrics are still changing over time.
For large-scale training jobs, it can take hours or even days for a silent error
to show obvious signs of anomalies in these metrics. Moreover, these metrics
can be highly noisy in real-world training tasks. Using them for detection can
easily create false alarms. When a silent error causes a performance impact, it can
also be difficult for developers to tell whether or not the long training iteration is
expected. Even after the error is detected, developers often have no
clear clues of the root cause. The diagnosis process is onerous and often based on ad-hoc trial-and-error changes (e.g., to hyperparameters) 
and rerunning
the training job to check if the error is fixed, leading to both developer productivity
loss and wasted resources.

\subsection{Case Studies}

We describe two representative cases from our study.
\paragraph*{BLOOM-176B training (DeepSpeed-1801)} \label{sec:ds_1801_case}
This is the example described in \autoref{sec:intro}, which occurred in
HuggingFace's training of BLOOM-176B in 2022~\cite{huggingface2022bloom}.
Due to the large model size, 3D parallelism~\cite{shoeybi2020megatronlmtrainingmultibillionparameter} was used to partition the model across multiple
GPUs and nodes. Conceptually, tensor parallelism partitions the model by
splitting individual layers into multiple GPUs. Thus the same layer on
different GPUs is not the same.  However, in implementation, due to the
communication cost, certain layers that do not impose a memory bottleneck are
not partitioned. In the Megatron-style tensor parallelism~\cite{shoeybi2020megatronlmtrainingmultibillionparameter},
LayerNorm layers are not partitioned as their size is small (<1\% of the model
weights) compared to the other layers, like attention layers.  Thus, optimizers
must be aware of this partition scheme, carefully distinguish between different
kinds of layers, and perform updates accordingly. This leads to complex
logic in the optimizer.

In this specific training task, DeepSpeed's BF16Optimizer was used.
It had a bug that causes the gradient clipping to be enabled on only the first
GPU on layers that are not partitioned. This caused the LayerNorm layers'
weights to silently diverge, as these layers were updated with different
gradients.

This bug neither triggers exceptions nor immediately affects loss or accuracy. 
Only when the model partitions across GPU need to be merged
into one checkpoint file will users realize that the model has
diverged. This happens in two scenarios: (1) when training completes, and the
model needs to be served or further fine-tuned; %
(2) during training, there is a need to change the parallelism configuration,
e.g., due to failure of a GPU. The developers of BLOOM-176B were lucky enough
to catch this error before the model was served in production or further fine-tuned 
while investigating another error. The detection took 10 days, and it required 9
additional days to merge the weights and mitigate the impact of divergence. 

We conduct an experiment to further confirm the impact of this error on a small
scale. Specifically, we train a small transformer-based language model using
the CodeParrot clean dataset \cite{codeparrot_clean_train}, with 4 tensor parallel
(TP) ranks and 2 data parallel (DP) ranks. The results are shown in \autoref{tab:bug-impact-ds-1801}.

At the initial steps (2000 and 4000) iterations, the
loss and the perplexity of the model on the validation and test set are both
noticeably affected by merging the weights, and the difference increases with the number
of training iterations.

\paragraph*{PyTorch-115607}
Another interesting case is PyTorch-115607. It showcases situations when
a silent training error can be detected, but the diagnosis is frustrating.
\inlinecode{torch.dynamo}\cite{Ansel_PyTorch_2_Faster_2024} is a new feature in
PyTorch that provides a Just-In-Time (JIT) compiler for unmodified PyTorch
programs. It transparently swaps bytecode to be executed with a more optimized
version and thus can speed up the training process. It uses guards to
ensure that when certain conditions change (e.g., the shape of a tensor changes), 
the bytecode is recompiled to reflect the changes. Ideally, every critical
variable that can affect training should be guarded. However, in this case, a
guard is missing, which causes the model to not update if a developer decides
to only do a forward pass without a backward pass in a certain iteration.
This error can be detected since the entire model is not updating. However,
diagnosing it can be problematic: if the training program does not dump
per-iteration loss and accuracy, the developer would have no idea where the
model stopped updating.

\subsection{Implications}
\label{subsec:implication}
Our study underscores the significant challenges posed by silent errors in DL
training. It also reveals implications for designing solutions to
address such errors.

Manual methods for detecting and diagnosing silent errors are time-consuming,
ineffective, and impractical, given the scale and complexity of the DL training
system stack. Automated approaches are essential to reduce the burden on
developers while improving the reliability of training pipelines.

Monitoring high-level signals such as loss and accuracy is insufficient. These
metrics are often noisy and periodic, leading to missed or delayed detection.
They are also prone to false alarms because it is difficult to distinguish
between expected fluctuations and true anomalies. Moreover, they provide few
clues for debugging. Fundamentally, they are not designed for error detection.
A true detection solution is required and needs to operate at a deeper level
than such signals.

The diverse locations and root causes of silent errors call for systematic
approaches. While point solutions, such as using differential
testing~\cite{NNSmith2023ASPLOS,pham_cradle_2019} to find bugs in DL
compilers and frameworks, have been proposed, they offer only limited help in
addressing the broad categories of silent errors.

Eliminating all bugs statically is ideal, but it is nearly impossible due to
the complexities involved in DL training. Many bugs tend to only trigger in a
specific environment, dataset, or scale.  Continuous monitoring of a DL
training task provides a safety net to detect silent errors at runtime.

\section{System Design}
\label{sec:design}
Motivated by our study, we design \sys, a framework that takes a proactive
validation approach to quickly and reliably detect silent DL training errors
with a wide variety of root causes, while providing diagnosis clues.

The designs of \sys are guided by two key insights we have from analyzing
real-world cases. First, while the symptoms of silent training errors take time
to become visible and often appear non-deterministic in generic, high-level
signals such as loss, the root causes of many errors are triggered early on and
detectable through specific, lower-level checks. Second, seemingly unrelated DL
training programs can share similar correctness properties due to the
heavy reliance on external libraries and similar training methods.

Based on these insights, \sys introduces the notion of \emph{training
invariants}, which are rules that should hold during training, \emph{e.g.}, the
model weights should stay consistent across distributed workers. These rules
capture the semantics and correctness properties specific to a training task.
Interestingly, we observe that a training invariant often only applies under
specific conditions, \emph{e.g.}, to particular layers or
parallelism settings, \emph{i.e.}, they have \emph{preconditions}.
\sys automatically infers training invariants as well as their
preconditions and enforces them to detect silent errors.

\textbf{Scope.} Silent errors in DL training span a broad spectrum, ranging from
correctness violations such as buggy API implementations to
optimization-sensitive issues like hyperparameter choices. This work
specifically focuses on correctness violations, which directly affect the
integrity and correctness of training outcomes, and are often high-impact and
actionable, yet remain largely overlooked by existing tools. Within this scope,
\sys aims to provide early and accurate \emph{detection} of silent training
errors before they silently propagate and accumulate. While its detection
results can offer useful diagnosis hints to developers, providing systematic
debugging support for precisely identifying root causes deserves dedicated
investigation in future work.

\subsection{Overview}

\begin{figure}[t]
    \centering
    \includegraphics[width=3.3in]{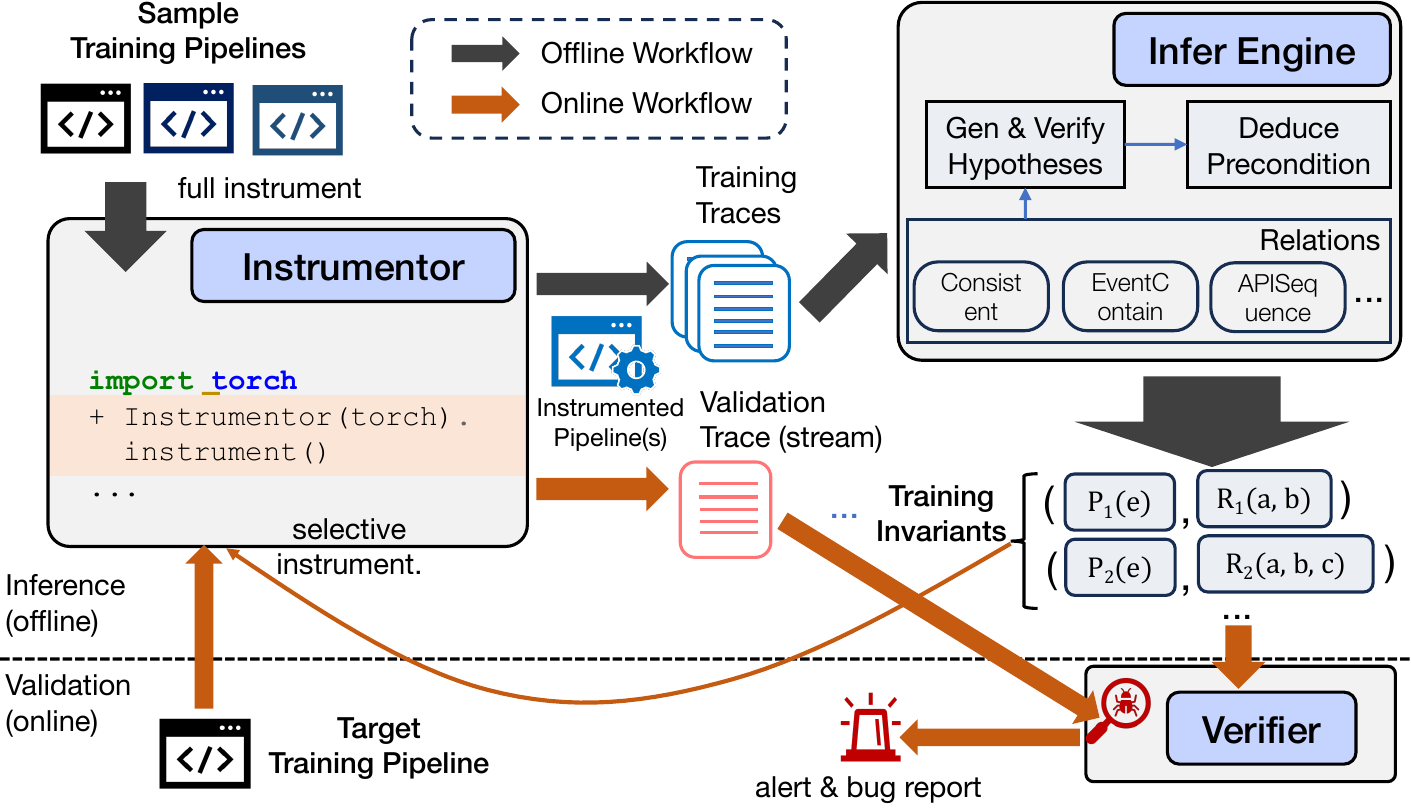}
    \caption{Workflow of \sys.}
    \label{fig:ml-daikon-workflow}
\end{figure}

\autoref{fig:ml-daikon-workflow} shows the workflow of \sys, which operates in
two phases. In the offline phase, it automatically infers a set of training
invariants from high-quality training pipelines. In the online phase, these
invariants are deployed to a given training job to check for violations
during training. \sys reports invariant violations with contextual
information to assist confirmation and investigation.

\sys comprises three major components: (1) \instrumentor, which
dynamically instruments DL training programs to collect runtime traces with low
overhead; (2) \inferengine, which analyzes these traces to infer
training invariants and their preconditions automatically; and (3)
\checker, which continuously validates a training task against the
invariants. When \instrumentor is used in the online stage, it performs
selective instrumentation relevant to the inferred invariants.

\begin{figure}[t]
    \centering
    \includegraphics[width=\columnwidth]{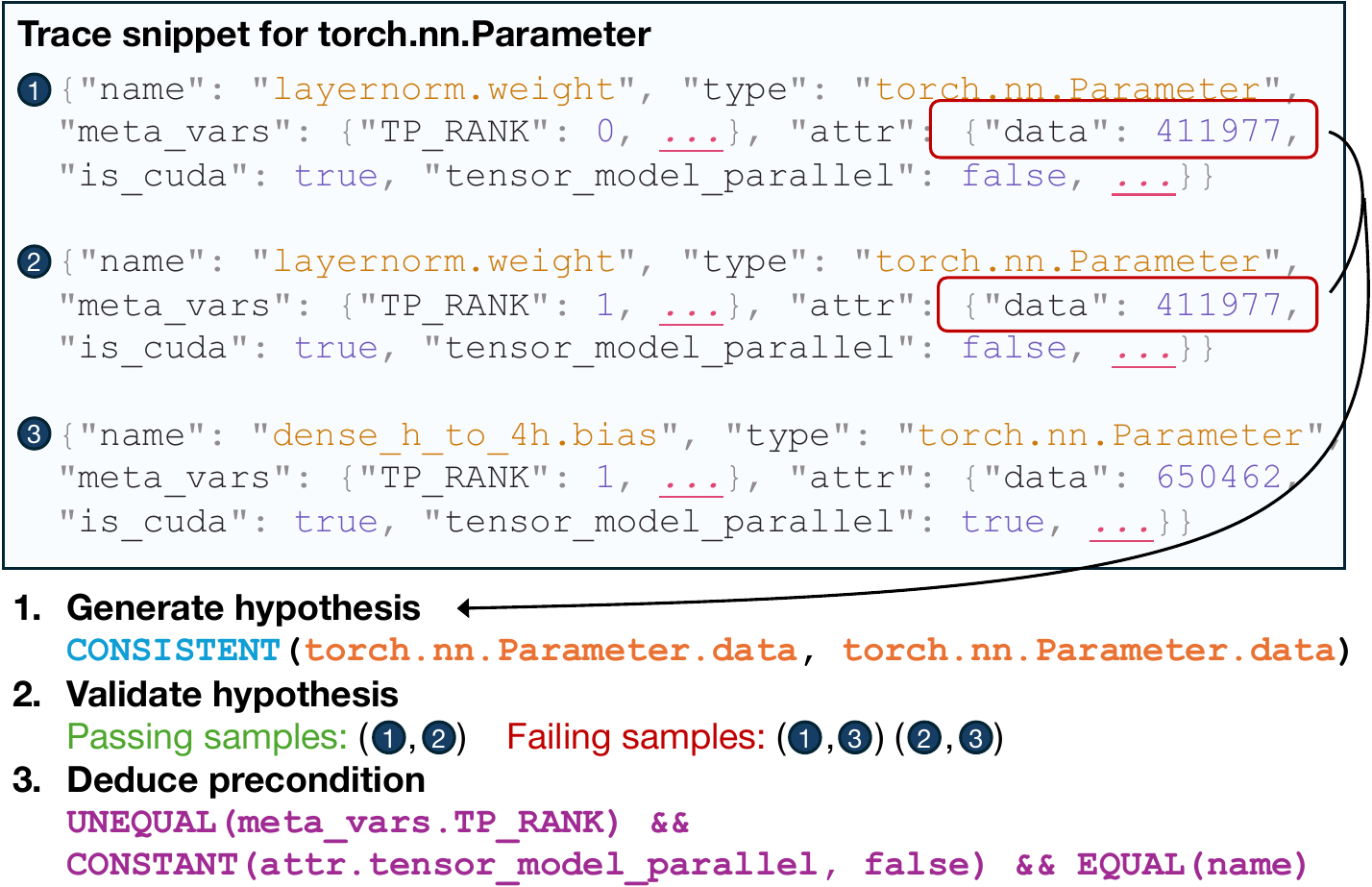}
    \caption{The invariant with preconditions \sys infers for
    Bloom-176B error (\cref{sec:ds_1801_case}), and the simplified
    trace it emits and uses.}
    \label{fig:trace-ds1801-simple-snippet}
\end{figure}

In this section, we focus on \inferengine. \autoref{sec:impl} will
describe the design of \instrumentor. \autoref{fig:trace-ds1801-simple-snippet}
shows a simplified example of the concrete training invariant with
preconditions \sys infers for the Bloom-176B training error.

\subsection{Invariant Representation}
\sys focuses on capturing semantic rules related to DL training.
Such rules can be expressed as logical relations involving key
variables (e.g., model weights), and events (e.g., API invocations).
\inferengine defines a generic \emph{relation interface} from which the specific
relations can be implemented. An invariant is a relation instantiated with concrete
variables and APIs that must satisfy the specified relationship.

\begin{table}[t]
  \setlength{\tabcolsep}{2pt}
  \centering
  \footnotesize
  \begin{tabular}{@{}p{3.2cm}p{5cm}@{}}
  \toprule
  \textbf{Relation} & {\textbf{Description}} \\
  \midrule
  $Consistent(V_a, V_b)$  & $V_a$ and $V_b$ should have the same values, while the values may change \\
  $EventContain(E_a, E_b)$ & $E_b$ must happen in the duration of $E_a$ \\
  $APISequence(I_a, I_b, \ldots)$ & $I_a$, $I_b$, $\ldots$ must all occur and in the specified order \\
  $APIArg(I_a, is\_distinct)$ & Ensures argument consistency or distinction in all calls to $I_a$\\
  $APIOutput(I_a, bound\_type)$ & The output of $I_a$ must meet certain attribute constraints \\
  \bottomrule
  \end{tabular}
  \caption{Sample relations \sys defines and supports.}
  \label{tab:relations}
\end{table}

Through analyzing silent training errors in practice, we identify and provide a
set of representative relations often violated in these silent errors. 
\autoref{tab:relations}
lists five relations \sys supports. For example,
\begin{itemize}[noitemsep, topsep=0pt, partopsep=0pt, leftmargin=*]
  \item \textbf{Consistent} relation establishes that two attributes of
    different variables should be equal despite how the value changes. This
    is effective for distributed training, where the model layers
    can be partitioned or duplicated across workers, and for model architectures
    that use shared parameters.
    \item \textbf{EventContain} relation establishes that when an API is invoked, a specified
    child event, such as another API call or a variable state change, occurs
    within that invocation. For example, calling \inlinecode{Optimizer.step}
    should always contain updates to the model parameters and the
    corresponding operations that perform those updates. This relation is
    effective for catching misconfigurations or unexpected arguments that
    otherwise lead to silent control-flow deviations.
  \item \textbf{APISequence} relation establishes (1) the group of APIs that
    should be called together and (2) the order in which they should be called.
    Users often forget to call certain APIs or call them in the wrong order,
    which can lead to silent errors. For example, in a few rookie mistakes
    collected from StackOverflow, users forgot to call
    \inlinecode{Optimizer.zero\_grad} before \inlinecode{loss.backward},
    leading to training instability.
\end{itemize}

These relations are deterministic and encode strict semantics, enabling precise
and timely detection of silent errors. Besides the built-in relations, \sys is
extensible, so developers can easily add new relations into the framework.

Invariant inference in \sys begins by instantiating each relation with concrete
\emph{descriptors}. A descriptor is a predicate that selects which trace records
or events an invariant should examine. API descriptors specify the API's name
and, optionally, the expected arguments or return values. Variable descriptors
specify a variable's type and attribute name, along with an optional expected
value or value change.

\subsection{Trace Representation}
\label{sec:trace-representation}

\sys collects execution traces from a DL training 
program by instrumenting it to emit relevant runtime information. 
A raw trace consists of a sequence of records capturing API entry and exit points, 
as well as variable states. Each trace record is annotated with a timestamp and thread ID.
\sys further extracts high-level \emph{events} from raw trace records to represent 
semantically meaningful behaviors. 
For example, an \emph{APICallEvent} represents a complete API invocation, 
aggregating its entry and exit records along with derived attributes such as execution duration and nested events. 
These high-level events form the foundation for invariant inference by providing structure to raw trace data.

\sys additionally introduces \textbf{meta variables}, contextual attributes for a trace record.
These include properties such as the training iteration number, distributed training ranks, 
and active context managers. Users can also define custom meta variables, 
such as pipeline stage (\emph{e.g.}, initialization, training, or evaluation).
These meta variables are essential for precondition inference (\autoref{sec:precondition}), 
enabling \sys to generate invariants that are both precise and interpretable.

\setlength{\algomargin}{1em}
\begin{algorithm}[t]
  \small
  \caption{Invariant Inference}
  \label{algo:invariant_inference}
  \KwIn{traces, relation\_pool}
  \KwOut{all\_invariants}
  all\_invariants $\gets$ []\;
  \ForEach{relation $\in$ relation\_pool}{
      hypotheses $\gets$ []\;

      \ForEach{trace $\in$ traces}{
          hypotheses.extend(relation.GEN\_HYPOS(trace))\;
      }

      \ForEach{hypo $\in$ hypotheses}{
          \ForEach{trace $\in$ traces}{
              relation.COLLECT\_EXAMPLES(trace, hypo)\;
          }
      }

      \ForEach{hypo $\in$ hypotheses}{
          preconditions $\gets$ INFER\_PRECONDITION(hypo)\;
          \If{preconditions $\neq$ null}{
              hypo.invariant.preconditions $\gets$ preconditions\;
              all\_invariants.append(hypo.invariant)\;
          }
      }
  }
  \Return{all\_invariants}\;
\end{algorithm}

\subsection{Invariant Inference}
\sys takes a hypothesis-based approach to infer training invariants from traces.
As summarized in \autoref{algo:invariant_inference}, the inference workflow 
contains three main steps. (1) \textbf{Hypothesis generation}: The engine scans 
through all traces to instantiate a relation with potential concrete descriptors. 
(2) \textbf{Hypothesis validation}: For each hypothetical invariant,
\sys validates it against the trace and records entities that match the
descriptors as passing/failing examples based on whether the relationship
holds. (3) \textbf{Precondition deduction}: Try to deduce the distinction between
passing/failing examples. Such distinction, if found, will be used as the
precondition for the property, and if not found, the hypothetical property is
invalidated and dumped. 

\begin{algorithm}[t]
  \small
  \caption{Hypothesis Generation for \texttt{Consistent}}
  \label{algo:consistency_gen}
  \SetAlgoLined
  \SetKwIF{If}{ElseIf}{Else}{if}{}{else if}{else}{end if}
  \KwIn{trace}
  \KwOut{Generated hypotheses}
  variables $\gets$ trace.get\_all\_variables()\;
  \ForEach{$(var_1, var_2) \in \text{Combinations}(variables, 2)$}{
      \ForEach{$(attr_1, attr_2) \in \text{CartesianProduct}(var_1.attrs, var_2.attrs)$}{
          \If{exists\_value\_match($attr_1.states, attr_2.states$)}{
              hypo $\gets$ NEW\_HYPO($relation \gets \text{ConsistentRelation}$,\\
                  \quad $entities \gets [$VarDesc($var_1.type, attr_1.name$),\\
                  \quad VarDesc($var_2.type, attr_2.name$)$]$)\;
              \textbf{yield} hypo\;
          }
      }
  }
\end{algorithm}

Each relation type implements the methods to generate and validate hypotheses
for that relation. These methods are invoked in the generic inference loop.
For the motivating example, the \inlinecode{Consistent} relation can be
instantiated with two \inlinecode{torch.nn.Parameter} object's
\inlinecode{data} attribute (\autoref{fig:trace-ds1801-simple-snippet}).
\autoref{algo:consistency_gen} shows how the hypotheses are generated
for this relation. The other relations' hypotheses generation
and validation are defined in a similar vein.

\subsection{Preconditions}
\label{sec:precondition}
Semantics in deep learning are often context-sensitive and can only be
applied to a specific subset of the training pipeline. For example, the
parameter consistency invariant in distributed training should only be checked
for the same model layer across workers within the same training iteration
and only for the parameters replicated instead of partitioned across
workers. For another example, the output tensor dtype usually depends on the
input tensor dtype, but when an autocast context manager is active, the output
tensor dtype should be the autocast dtype instead of the input dtype.

\sys thus defines \emph{preconditions} for a training invariant. Preconditions
provide various benefits: (1) confidence in the invariant's accuracy, (2)
reduction in false positives, (3) debugging, as when the invariant is violated,
the precondition can help explain which part of the pipeline is causing the
error, and (4) reduced overhead by only performing the applicable checks. In
addition, preconditions enable \textbf{transferable invariants}, as they
provide a very clear context of when the invariant should be applied and thus
can be applied to other pipelines that share the same context.

\subsection{Deducing Preconditions}
\label{paragraph:deduce-preconditions}
We design an algorithm to deduce the weakest yet safe preconditions for each
invariant.  A precondition is considered \emph{safe} if it provides a clean
separation: evaluating to \texttt{true} for \emph{all} passing examples and
\texttt{false} for \emph{all} failing examples.

A precondition consists of one or more \emph{conditions} (predicates) that
compare a field's values across all trace records of a given example. \sys
supports four types: \texttt{CONSTANT}, where the field's value is identical in
every record and must match a specific required value; \texttt{CONSISTENT},
where the field's value is identical in every record without a specific value
constraint; \texttt{UNEQUAL}, where the field takes different values across
records; and \texttt{EXIST}, where the field appears in every record.

The deduction algorithm first scans through the passing examples and produces
conditions for each example. It then forms a \emph{candidate precondition} by
using the conditions found. For each example, the algorithm finds the
conditions satisfied in all trace records of that example.  The candidate
precondition is then formed by taking the conjunction of the conditions
satisfied in all passing examples. This conjunction is verified
against the failing examples to see whether the precondition is safe; if so,
the algorithm returns that precondition.

By restricting the candidate precondition to a conjunction of conditions that
hold in all passing examples, the algorithm implicitly assumes the invariant
applies to only a single scenario, which is often not the case. For instance,
the parameter consistency invariant holds both across (1) data-parallel workers
and (2) LayerNorm parameters on tensor-parallel workers.When the initial
candidate precondition is unsafe, the algorithm attempts to split the passing
examples into subgroups based on remaining non-overlapping conditions and
performs inference on each subgroup. If no further splitting is possible, the
algorithm terminates and reports an inference failure. If the preconditions
inferred from these subgroups are safe and collectively cover all passing
examples, they are combined disjunctively to form the final precondition.

\paragraph{Prune Irrelevant Conditions}
Numerous conditions can be inferred from the trace, but not all are relevant to
the invariant. For example, in the parameter consistency invariant for
distributed training, it is likely that all the parameters have the same
\inlinecode{is\_cuda} attribute.  \sys prunes irrelevant conditions during the
safety verification process by removing conditions not violated in any
failing examples. These conditions evaluate to true in all examples and are
not discriminative. This approach is simple yet effective in removing a good
amount of noisy conditions unrelated to the invariant.

A more complex situation is when the irrelevant condition is an artifact of the
invariant handled by the current algorithm.
For example, all consistent model weights will also have the same gradient
values to maintain consistency. However, using consistent gradient values as a condition for the
parameter consistency invariant is not a good idea because, while safe, it is
too shallow and prevents the algorithm from going deeper to find the multiple
scenarios that the invariant holds. To fix this problem, we allow each relation
to encode rules on what conditions should be avoided in the precondition
inference process. For example, a \texttt{Consistent} invariant, if about
attributes of type \texttt{torch.Tensor}, cannot use any other attributes of
type \texttt{torch.Tensor} as a condition. Alternatively, static analysis can
help determine correlated fields and remove them from the precondition
inference process, but its complexity and overhead may not be justified by
the benefits.

\paragraph{Example}
In the simplified example presented in~\autoref{fig:trace-ds1801-simple-snippet},
there is one passing example
and two failing examples.  Each example contains two trace records of
\inlinecode{torch.nn.Parameter} objects.  The algorithm first generates the
candidate precondition by finding the conditions that hold in the
positive example, which is \inlinecode{CONSTANT(tensor\_model\_parallel,
False) \&\& CONSTANT(is\_cuda, True) \&\& UNEQUAL(meta\_vars.TP\_RANK)}

Verifying the conjunction of these three conditions against the failing
examples reveals that it is safe. However, the second condition of
\inlinecode{is\_cuda} constantly \inlinecode{True} is not violated
in any failing examples; thus, it is pruned from the precondition. The final
precondition is \inlinecode{CONSTANT(tensor\_model\_parallel, False) \&\&
UNEQUAL(meta\_vars.TP\_RANK)}

When the candidate precondition is unsafe, we are in an
\emph{under-constrained} situation. This can happen for three reasons: (1) the
precondition is fundamentally not expressible in the condition types we
support, (2) the information in the trace is not enough to infer the
precondition, or (3) the invariant holds under multiple preconditions. In our
experience, (3) happens most of the time. This indicates that local attributes
and meta variables are sufficient for most invariants and do not need 
a complex grammar to express the preconditions. 

\begin{figure}
    \centering
    \includegraphics[width=2.5in]{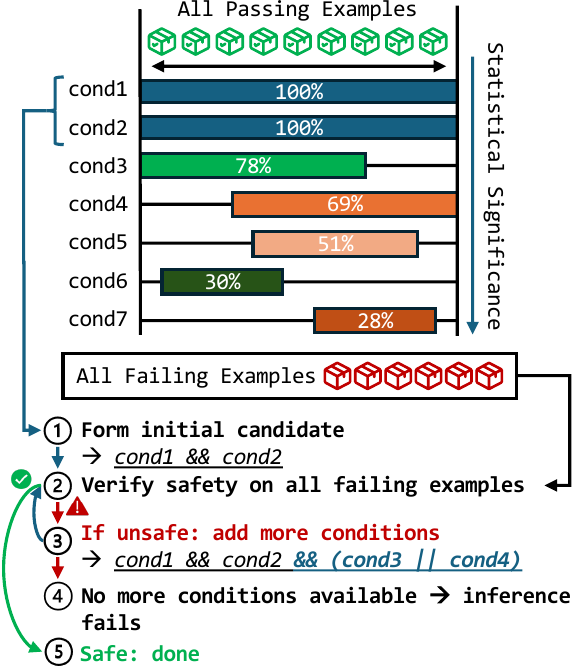}
    \caption{Deduce precondition in under-constrained situations.}
    \label{fig:precond-infer-process-exp}
\end{figure}

We enhance an unsafe candidate precondition by adding conditions not considered
previously to further constraint it, as described in \autoref{paragraph:deduce-preconditions}.
We choose the conditions to add based on decreasing order of
statistical significance, i.e.,  the conditions that cover the most passing
examples.  In~\autoref{fig:precond-infer-process-exp}, the candidate
precondition is the conjunction of the two fully-covering conditions,
\texttt{cond1 \&\& cond2}.  However, since it is unsafe, we attempt to
add \texttt{cond3} and \inlinecode{cond4}, resulting in a safe precondition of
\inlinecode{cond1 \&\& cond2 \&\& (cond3 || cond4)}. If this precondition is
still unsafe, we repeat the process until we find a safe precondition, or the
computation budget is exhausted.  The statistical significance based search
helps reduce the search space and find the majority scenarios that the
invariant holds.

Our algorithm is not guaranteed to find strictly the weakest precondition, as
the pruning strategy only looks at individual conditions, and we do not rely on
any static program analysis to infer the preconditions. However, it provides
a good trade-off between simplicity and effectiveness.

\subsection{Filtering Out Superficial Invariants}

One challenge in invariant inference is distinguishing meaningful invariants
from superficial ones---those that appear valid due to the limited information
available in a trace. 
For example, two irrelevant APIs \texttt{torch.cuda.is\_available()} and
\texttt{torch.jit.is\_scripting()} can have consistent return values.

\sys deems an invariant superficial if it cannot deduce a precondition for this
invariant. Such invariants will not be deployed because they may not hold in
the target training task. Even if they hold, they are not effective as runtime
checks because we do not know when to apply them due to the
lack of precondition. This is a key design choice to reduce false
positives and enhance explainability.

Our approach differs significantly from traditional invariant mining solutions.
They typically use the number of passing examples and the number of failing
examples to estimate the statistical significance and select likely invariants.
This approach has two drawbacks in our context. In DL training, having a large
number of failing examples does not mean the invariant is superficial due to
the diverse and intricate training process. Local semantics might not
be statistically representative in the global context. For example, in tensor
parallelism, all major parameters in the attention layer are partitioned across
workers. Only the LayerNorm parameters are replicated, which accounts for less
than 0.1\% of the total parameters in representative transformer models like
GPT \cite{Radford2019LanguageMA}. During inference, the invariant
that \texttt{torch.nn.Parameter} objects should be consistent has a
passing/failing ratio of 1:38. This can be pruned by a statistical significance-based 
approach, but in reality, this turns out to be an important
invariant for catching the Bloom-176B error.

\subsection{Scalability}

DL training produces large traces due to frequent API calls and variable
updates, posing significant challenges for analysis. For example, instrumenting
a 2-GPU, 70M-parameter pretraining run for BLOOM-176B (using
Megatron-DeepSpeed's GPT-2 pipeline) generates approximately 92,000 trace
records (50\,MB) per training iteration.

\sys adopts two key design choices to address this challenge. First, it
restricts analysis to a predefined set of high-level semantic relations,
significantly reducing the search space. Second, \sys abstracts APIs and
variables into \emph{descriptors}. For variables, they are typically
compound objects with multiple attributes. A variable descriptor consists of the
variable's type (as returned by \inlinecode{type(obj)}), attribute name (\emph{e.g.},
\texttt{data}, \texttt{grad}), and optional value constraints. This allows \sys
to reason over groups of variables sharing the same type and attribute, rather
than enumerating individual instances. For example, when analyzing BLOOM-176B
using a 2-GPU and 70M-parameter run, enumerating 104 variable instances such as
\texttt{0.input\_layernorm.weight} and
\texttt{2.post\_attention\_layernorm.weight} would yield 5,356 pairs to
consider. In contrast, \sys considers only the available PyTorch variable types
relevant to training state, where the primary type is
\inlinecode{torch.nn.Parameter}. This dramatically reduces the number of
invariant candidates while preserving the properties needed for error
detection. Descriptors can also encode constraints (\emph{e.g.}, requiring
non-null values) to further generalize invariant specification.

\subsection{Input Requirements}
Importantly, while the inferred invariants are intended to validate large-scale
training, generating them does not always require large-scale setups. In our
experiments, effective invariants can often be inferred from small-scale runs.
For example, although the original BLOOM training job spanned hundreds of GPUs,
\sys was able to infer the relevant invariant using only a 2-GPU run. All
evaluated invariants in this paper were inferred from training jobs using at
most 4 GPUs and 100 iterations. These results demonstrate that representative
behaviors can be captured with lightweight workloads, making the inference phase
practical and efficient.

\section{Implementation}
\label{sec:impl}

We implement \tool in Python with \SI{22.7}{K} lines of code. The system is
composed of \instrumentor, \inferengine, and \checker. A key challenge in
developing \sys is to balance fine-grained trace collection for
inferring effective invariants to detect silent errors and minimizing
runtime overhead. We made extensive efforts to explore different techniques
for achieving a good trade-off.

\subsection{\instrumentor}
\instrumentor collects traces from a training task for both \inferengine and
\checker. It instruments a given DL training program to emit information as
required by the relations we support (\autoref{sec:design}), specifically (1)
\emph{API invocation trace:} function entry, exit, arguments, and return
values, (2) \emph{variable state trace:} variable assignment, deletion, and
modification, and (3) \emph{meta variables:} steps, epochs, ranks, etc.

We implement it as a command-line tool that takes the path to the
entry Python program and the libraries of interest (\emph{e.g.}, \texttt{torch},
\texttt{deepspeed}) as arguments. \instrumentor automatically scans the \texttt{import}
statements and model definitions in the program and instruments the necessary
functions and variables.  The user can also provide a shell script that sets up
the environment and arguments for the program. Trace logs are written to a
specified directory using JSON format.

\instrumentor is designed with three key goals: (1) non-intrusiveness, (2) low overhead, and (3) high coverage.
\instrumentor also aims to be flexible in instrumentation granularity, as different relations require different levels of
detail. For example, the \texttt{Consistent} relation only requires a periodic
sampling of model states, whereas the \texttt{EventContain} relation
requires an eager model state logging whenever the variable is modified, as
accurate timing of events is crucial for it.

One option is to use Python's \inlinecode{sys.settrace} function, which installs
a trace function that is invoked on every function call, return, and exception.
However, this approach incurs prohibitively high overhead; we observed a
200$\times$ to 550$\times$ slowdown. It can also interfere with existing
workflows that use tools relying on \inlinecode{sys.settrace}, such as
\inlinecode{pdb} and \inlinecode{cProfile}.

\paragraph{Dynamic Monkey Patching}
To meet the above goals, we adopt a \emph{monkey-patching} approach.
\instrumentor is implemented as a Python package that dynamically instruments
the target program by injecting hooks into relevant source code at runtime. It
supports \emph{selective} instrumentation, allowing users to specify which
modules to instrument; only APIs defined in the specified modules are patched.
During instrumentation, \instrumentor recursively traverses the namespace of
each selected module and wraps identified methods. Each wrapper inserts logging
and bookkeeping logic before and after invoking the original function. To
minimize overhead, \instrumentor skips low-level internal functions,
particularly those in \inlinecode{torch.jit} and \inlinecode{torch.\_C}, which
are invoked frequently but rarely provide meaningful information for invariant
inference.

\paragraph{Tracking Variables}
\label{sec:proxying}

Instrumenting Python variables is more challenging than tracing APIs.
In CPython, assignment operations occur at the C level and do not invoke any
Python-level hooks, making it infeasible to efficiently track arbitrary variable
state changes. Our study shows that most non-trivial silent errors, despite their
diverse root causes, affect a small set of key objects, such as the model and
optimizer. If a silent error does not impact model quality, it is often
inconsequential for training correctness. This leads to a key insight: tracking
the state of only the model and optimizer is sufficient to detect meaningful
silent errors. These objects are typically long-lived, and updates to them occur
through attribute modifications rather than object replacement, which simplifies
tracking. Therefore, \instrumentor focuses on tracking models and optimizers
rather than arbitrary local variables.

To achieve this, models and optimizers are wrapped with a \inlinecode{Proxy}
that intercepts state-changing operations via overridden magic methods such as
\inlinecode{\_\_setattr\_\_}. These changes are logged to the trace eagerly upon
execution. During instrumentation, \instrumentor scans the program’s AST to
locate the initialization sites of these objects and replaces them with their
corresponding \inlinecode{Proxy} instances. Alternatively, when precise timing
of state changes is not required, \instrumentor supports a lower-overhead,
sampling-based approach that registers a state-dump callback on
\inlinecode{Optimizer.step}.

\paragraph{Logging Hashes of Tensors} When dumping the model states, we are essentially doing
checkpointing. The cost of this is unbearable as the model states are large and
the overhead of serializing them, writing them to the log file, and reading
them back is significant.  Through our study, we observe that the actual values
of the tensors are typically unimportant for inferring invariants. Silent errors
typically arise from the shape, dtype, and the equality relationship between
the tensors. Thus, \instrumentor only logs the hash of tensors.

\paragraph{Collecting Meta Variables} \instrumentor also collects meta
variables such as the current step, epoch, rank, \emph{etc.} Whenever a trace record
for a function call or variable state is dumped, \instrumentor walks through
the call stack and finds the loop index local variable. This is a simple heuristic
that works for most cases, as the loop index is usually a local variable in the
outermost loop that is incremented in every iteration.  \instrumentor further
allows users to specify meta variables by calling the \inlinecode{set\_meta}
API. Users may also annotate the program into different phases, such as
training, validation, and testing. The general idea for collecting meta
variables is to provide context for the invariants to deduce preconditions.

\subsection{\inferengine}
\inferengine processes the trace files generated by \instrumentor to infer
training invariants and their preconditions. One key challenge is handling
large input traces produced by the data-intensive nature of DL training.
Typical training pipelines can generate several hundred megabytes of trace data
per epoch. Our algorithms, as described in \autoref{sec:design}, are designed
to efficiently deduce invariants with preconditions from large traces.

We implement \inferengine in Python to leverage its rich ecosystem of libraries
for data processing and deep learning. In addition, our traces cannot be
conveniently represented with a fixed schema. We implement multiple trace
backends, including Pandas, Polars, and built-in dictionaries. From
experimenting with these backends, we choose Pandas DataFrames with Python
dynamic typing as the default based on its analysis performance and
flexibility. We introduce optimizations with custom analysis functions that 
provide query caching, sampling, and pruning (\emph{e.g.}, pruning 
candidates related to \texttt{torch.cuda.is\_available}).

\subsection{\checker}
During the online checking phase, \tool consumes the stream of the trace generated
by the instrumented program to detect silent errors in real-time.
Different from the offline phase, the instrumentation is restrained to only the
APIs and variables that are relevant to the deployed invariants and is thus
lightweight. \checker monitors the trace and triggers a check when a
relevant piece of trace is available. It first evaluates whether the preconditions of an
invariant are satisfied. If so, it checks whether the invariant holds.
When an invariant violation is detected, \checker reports the invariant and the corresponding
trace to provide debugging help to developers.

\section{Evaluation}
\label{sec:eval}

We evaluate \sys to answer several questions: (1) Can \sys infer effective
training invariants for detecting real silent errors? (2) How quickly
can the invariants detect the errors? (3) Can \sys help diagnose a detected
silent training error? (4) Is the detection accurate? (5) What is
the runtime overhead?

Our experiments are run on a server running Ubuntu 22.04, equipped with an
Intel(R) Xeon(R) Silver 4310 CPU, 252 GB RAM, and one NVIDIA A40 GPU.
For trace collection related to distributed training, we use another
server with identical specifications but featuring 8 NVIDIA A2 GPUs. We use
Python 3.10 and PyTorch 2.2.2 with CUDA 12.1.

\begin{figure}[t]
  \begin{subfigure}[b]{0.35\linewidth}
    \centering
    \includegraphics[width=\linewidth]{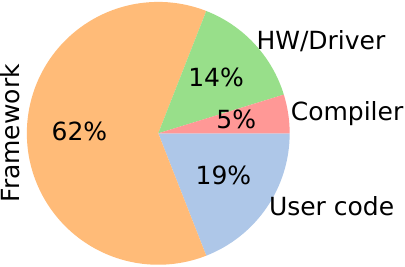}
    \caption{Error locations.}
    \label{fig:evalset:location}
  \end{subfigure}%
  ~
  \begin{subfigure}[b]{0.63\linewidth}
    \centering
    \includegraphics[width=\linewidth]{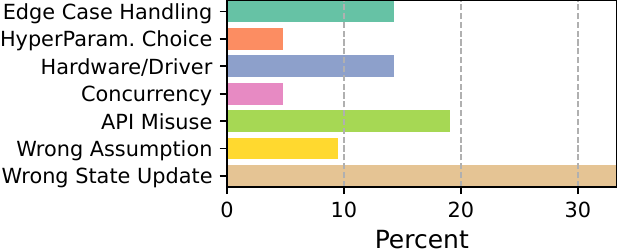}
    \caption{Root cause types.}
    \label{fig:evalset:cause}
  \end{subfigure}
  \caption{Root cause types of errors we reproduce for evaluation.}
\end{figure}

\begin{table*}[t]
    \centering
    \footnotesize
    \begin{tabular}{@{}lp{5.6in}@{}}
      \toprule
        {\bf Error Id} & {\bf Synopsis}\\
      \midrule
      B1 (PT-F84911)   & Misconfiguration that only BatchNorm layers are enabled causing training to not progress. \\
      B2 (PT-84803)    & Incorrect I/O setup causing data loss during P2P communication between GPUs. \\
      B3 (DS-1801)     & Logic bug in gradient clipping of BF16Optimizer causing model weights for LayerNorm layers to go out of sync. \\
      B4 (SO-60335387) & Missed zero\_grad() call causing gradients to accumulate and training to be unstable. \\ 
      B5 (TF-23723)    & Logic bug in truncation of input causing model output to be longer than threshold. \\
      B6 (SO-67180955) & Data loader workers having the same seed causing identical data fetched by workers. \\
      \midrule
      B7 (LT-725)      & Incorrect function transform causing dispatcher to not respect autocast contexts and use wrong datatype for training. \\ 
      B8 (PT-115607)   & Missing guards causing jit to not recompile even when the code has changed, causing model to not learn. \\ 
      B9 (PT-104336)   & Module.to not preserve the gradient syncing hooks during serialization, causing DDP model weights to go out of sync. \\ 
      B10 (X-DDP)      & User incorrectly invoking forward on internal modules of DDP instead of DDP itself, causing DDP model weights to go out of sync. \\ 
      B11 (TF-34204)   & Input processor of Pixtral model only returns the first item when input data contains multiple items, causing significant amount of data not used during training. \\ 
      B12 (TF-33844)   & Flash attention implementations of models in transformers didn't set dropout to 0 during evaluation, causing unstable and non-reproducible evaluation results. \\ 
      B13 (TF-17877)   & Logic bug in distinguishing initialized/uninitialized parameters causing repeated initialization, leading to performance issues and unknown accuracy impact. \\ 
      B14 (TF-29903)   & Certain LSTM weights are not solved due to logic bug in the checkpointing function. \\ 
      B15 (TF-33455)   & Logic bug in calculation of total training steps when gradient accumulation is enabled cause training to stop earlier than expected. \\ 
      B16 (DS-5794)    & Logic bug in MoE layer implementation causing all2all to be invoked with conflicting arguments, leading to training hang. \\ 
      B17 (PT-96600)   & Incorrect I/O setup causing data loss during P2P communication between GPUs. \\ 
      B18 (BC-86)      & Faulty driver causing data loss during P2P communication between GPUs. \\ 
      B19 (PT-51800)   & SpectralNorm layers require train-time forward to initialize weights; Running evaluation before any training leads to very misleading results. \\
      B20 (MM-702)     & Missed call to freeze\_stage inside the init method of SwinTransformer causes loss computation to not respect all computations during forward. \\ 
      \bottomrule
    \end{tabular}
    \caption{Silent training errors we reproduced for evaluation. Only the first 6 errors are from our study.}
    \label{tab:eval_known_failures}
\end{table*}

\subsection{Silent Error Detection} \label{sec:eval:issue-detection}
To assess \sys's effectiveness, we collect and reproduce \numofevalsetbugs
\emph{real-world} silent errors. Of these, 6 are drawn from our prior study
(\autoref{sec:ml-silent-issues}), while the remaining 14 are newly gathered from
GitHub issues for popular libraries (PyTorch, DeepSpeed, etc.),
StackOverflow, and social media. These errors span a broad range of root cause
locations and types, as illustrated in \autoref{fig:evalset:location} and
\autoref{fig:evalset:cause}. The complete descriptions of these errors are listed in \autoref{tab:eval_known_failures}.

\begin{table*}[t]
    \centering
    \footnotesize
    \setlength{\tabcolsep}{3pt}
    \begin{tabular}{l*{21}{c}}
    \toprule
        & B1 & B2 & B3 & B4 & B5 & B6 & B7 & B8 & B9 & B10 & B11 & B12 & B13 & B14 & B15 & B16 & B17 & B18 & B19 & B20 \\ 
    \midrule
    & \multicolumn{20}{c}{Detection} \\
    \midrule          %
    \tool             & 1         & 0         & 1         & 0         & 0         & 0         & 0         & 1         & 1         & 1         &  0         & 0         & 0         & \ding{54} & \ding{54} & 0         & 0         & 0         & 0         & 0         \\ 
    Spike + Trend     & 4 epo     & \ding{54} & \ding{54} & \ding{54} & \ding{54} & \ding{54} & \ding{54} & 4 epo     & \ding{54} & \ding{54} &  \ding{54} & \ding{54} & \ding{54} & \ding{54} & \ding{54} & \ding{54} & \ding{54} & \ding{54} & \ding{54} & \ding{54} \\
    Anomaly Detection & \ding{54} & \ding{54} & \ding{54} & \ding{54} & \ding{54} & \ding{54} & \ding{54} & \ding{54} & \ding{54} & \ding{54} &  \ding{54} & \ding{54} & \ding{54} & \ding{54} & \ding{54} & \ding{54} & \ding{54} & \ding{54} & \ding{54} & \ding{54} \\
    NeuRI/PyTea       & \ding{54} & \ding{54} & \ding{54} & \ding{54} & \ding{54} & \ding{54} & \ding{54} & \ding{54} & \ding{54} & \ding{54} &  0         & \ding{54} & \ding{54} & \ding{54} & \ding{54} & \ding{54} & \ding{54} & \ding{54} & \ding{54} & \ding{54} \\
    \midrule
    & \multicolumn{20}{c}{Localization} \\
    \midrule          %
    \tool             & \ding{212} & \ding{74} & \ding{212} & \ding{74} & \ding{74} & \ding{212} & \ding{74} & \ding{212} & \ding{212} & \ding{212} & \ding{74} & \ding{74} & \ding{74} & n/a  & n/a       & \ding{212} & \ding{74} & \ding{74} & \ding{212} & \ding{74 } \\
    Spike + Trend     & \ding{110} & n/a       & n/a        & n/a       & n/a       & n/a        & n/a       & \ding{110} & n/a        & n/a        & n/a       & n/a       & n/a       & n/a  & n/a       & n/a        & n/a       & n/a       & n/a        & n/a        \\
    NeuRI/PyTea       & n/a        & n/a       & n/a        & n/a       & n/a       & n/a        & n/a       & n/a        & n/a        & n/a        & \ding{74} & n/a       & n/a       & n/a  & n/a       & n/a        & n/a       & n/a       & n/a        & n/a        \\    
    \bottomrule
    \end{tabular}
    \caption{Evaluation of detection time in terms of number of steps since the root cause is triggered and root cause localization for \tool and baseline detectors. 
    epo: number of steps in a complete epoch, \ding{54}: errors not detected at all. \ding{74}: capture the root cause. \ding{212}: detect on the effect chain. \ding{110}: no information about the root cause.}
    \label{tab:detection-timeliness-and-localization}
\end{table*}

We compare \sys with four baselines that represent the current practices and state-of-the-art
research:
\begin{itemize}[noitemsep, topsep=0pt, partopsep=0pt, leftmargin=*]
  \item \emph{Spike} detector is used to monitor numerical instability in
    training where the loss or accuracy spikes to a large value.
  \item \emph{Trend} detector is used to monitor the training process where the
    loss or accuracy is not decreasing or increasing as expected. A tolerance
    factor is set to allow some fluctuation.
  \item \emph{Anomaly Detection} detector applies common algorithms
    like LOF, Isolation Forest, and Z Score to detect anomalies on the same
    high-level metrics like loss and accuracy.
  \item \emph{PyTea\cite{PyTea2022ICSE}} and \emph{NeuRI\cite{liu2023neuridiversifyingdnngeneration}} are
    two recent research artifacts. PyTea specifies constraints on APIs used in the training process,
    primarily focusing on the shaping constraint of the input and output tensors.
    NeuRI automatically infers such constraints encoded in PyTea's syntax.
\end{itemize}

When implementing the first three detectors, we monitor signals as per industry
practice \cite{wandb} and apply the same configuration parameters to all 
errors for a fair and consistent comparison.  For the spike detector,
we set the threshold to 75, and for the trend detector, we set the tolerance to
3. The number of neighbors for LOF is set to 2, and the contamination factor is
set to 0.1 for Isolation Forest. Other parameters use defaults provided by
Scipy and Scikit-learn.

\paragraph{Methodology}
We prepare a reproduction script for each error and run them to emit (1) runtime trace for the
checking of \sys and PyTea/NeuRI, and (2) high-level metrics (loss, accuracy,
gradient norm) for the anomaly detection-based detectors. We then run the
detectors on the traces and metrics and collect the detection results.
The invariants used by \sys are inferred from PyTorch's official GCN,
Autocast, and DDP examples, for PyTorch-related errors, and Megatron-DeepSpeed's
official GPT pretraining examples for DeepSpeed-specific errors, and the
official Transformers trainer examples for Transformers-specific errors.

We focus on \emph{true detections} (true positives) to avoid rewarding
detectors that indiscriminately raise many alarms. This ensures the result
reflects real error-detection capability. For example, we observe that
when using the anomaly detection detector, no matter how we tune it, it
raises numerous alarms throughout the training process, \emph{e.g.},
since the loss is dropping fast. To objectively determine the true positive, we
run the fixed versions of each error and check if the detector also raises
alarms in the error-free traces.

\paragraph{Detection}
\label{sec:eval:results-detection}
\sys successfully detects \numofevalsetbugsdetected out of the
\numofevalsetbugs errors. \autoref{tab:detection-timeliness-and-localization} shows
the detection result for each case. The invariants \sys infers and uses represent all five
relations in \autoref{tab:relations}. In all cases, detection occurs no later
than one iteration after the root cause is triggered.  For the motivating
example, the incorrect gradient clipping logic is triggered in the second
training iteration, and \sys shortly detects it in the third iteration. Despite
diverse root causes, \sys achieves high detection coverage and timeliness.  We
attribute its effectiveness to our invariant checking approach and the
precision of the inferred invariants.

\sys fails to detect two errors: TF-33455 and TF-29903. TF-33455 involves the
trainer stopping early due to an incorrectly calculated total number of training
steps, while the training process itself is correct. Detecting this error
would require monitoring the computed training steps and comparing them with the
intended arguments. Currently, \sys does not support tracking Python primitive
variables, as doing so would incur prohibitive overhead and require
modifications to the Python runtime. TF-29903 concerns a bug caused by a corrupted state dict constructed within the
\texttt{safe\_checkpoint} function. \sys fails to detect this case because (1)
this error is confined to the checkpoint function and does not impact the main
training logic, and (2) \sys does not analyze local variables.

In comparison, the signal-based detectors collectively only detect 2 errors,
which are extreme cases where the model stops learning entirely and the loss is
constant over epochs. The PyTea/NeuRI detector detects 1 error, which is from 
a bug in the transformers library where the processed data does not have the same batch size as
the argument, falling into the shaping constraints supported by PyTea/NeuRI.

\paragraph{Diagnosis}
While diagnosis is not the primary goal of \sys, we conduct analysis to
understand whether invariant violations can aid in debugging.  Among the 18
cases detected by \sys, the violation reports can pinpoint the exact root cause
in 10 cases and localize close to the root causes in 8 cases.  The diagnosis
hints for the one error detected by PyTea/NeuRI are on par with \sys.

\paragraph{Case Study: B1}
We introduced the issue Bloom-176B as a motivating example for this project.
The issue arises when model weights become inconsistent across workers due to a
logic error in the gradient clipping feature of DeepSpeed's newly introduced BF16 optimizer.

To address this, we inferred invariants from the official Megatron-DeepSpeed pretraining
GPT example configured with 3D parallelism in FP16. While API-level invariants were not directly
transferable---since the FP16 optimizer integrates gradient clipping into its logic,
whereas the BF16 optimizer implements it as a separate API---we successfully inferred variable
consistency invariants that captured the issue. Specifically, the inferred invariant states that
model weights should remain consistent across all workers under two conditions: (1) consistency
across data parallel (DP) ranks, and (2) consistency across tensor parallel (TP) ranks, with the
attribute \texttt{tensor\_model\_parallel} set to \texttt{False}.
Condition (1) captures data parallelism, while condition (2) captures tensor parallelism.

During runtime checks, violations of model weight consistency were detected only under condition
(2) and not condition (1). Furthermore, since only LayerNorm weights satisfy the condition of
having \texttt{tensor\_model\_parallel = False}, the invariant enabled developers to pinpoint the
handling of LayerNorm weights in the BF16 optimizer as the source of the issue.

In terms of detection timeliness, the issue can be identified in the first iteration after gradient
clipping is triggered. In our reproduction script, the issue was detected in the third training iteration,
immediately following the activation of gradient clipping in the second iteration. This prompt detection
is attributed to the invariant's enforcement of strict weight consistency, where even minimal discrepancies
(e.g., a divergence of 1e-8 in a single tensor value) trigger a violation report. Such subtle inconsistencies
 would be undetectable using signal-monitoring-based detection methods, underscoring the effectiveness of our approach.

\paragraph{Case Study: B11}
B11 is a issue in the input processor for the Pixtral model,
a newly introduced model in the Transformers library.
When provided with an input tensor of random batch size, the processor erroneously
returns only the first sample in the entire batch,
effectively reducing the model's training batch size to 1 and causing it to train on incomplete data.

\tool detects this issue by enforcing input tensor shape consistency
across the arguments and return values of the \texttt{processor.\_\_call\_\_} API,
as specified by its inferred invariant.
The violation report highlights a mismatch between the input tensor shape and the return value shape,
directly pinpointing the input processor's implementation as the root cause.
This invariant, inferred from the official Transformers visual-language model (VLM) fine-tuning example using
QWen-VL (which also utilizes a processor instead of a tokenizer), guided the developer to locate and resolve the issue efficiently.

In summary, \tool detects issues with diverse root causes with high precision and timeliness,
typically within a single training iteration.
This is enabled by its highly precise invariants and the iterative nature of training,
where issues are captured as their effects propagate by the end of the iteration,
even if the root cause is not immediately evident.

\tool's violation reports are fine-grained, focusing on low-level behaviors, making them clear,
actionable, and closely aligned with the root causes of issues.
In contrast, traditional signal-based approaches can identify the presence of an issue but often
fail to pinpoint its root cause. Instabilities in high-level metrics, such as those caused by bad
hyperparameters, initialization, data, incorrect implementations, or hardware issues, may not always
indicate a code bug and can even occur in non-buggy pipelines. As a result, signal-based detectors often
produce ambiguous or misleading reports with low confidence, making them less actionable.

\begin{table}[t]
    \centering
    \footnotesize
    \begin{tabular}{@{}lp{2.8in}@{}}
      \toprule
        {\bf Bug Id} & {\bf Synopsis}\\
      \midrule
        AC-2665 & Initializing the optimizer prior to wrapping the model with DDP causes training to not progress. \\
        DS-6770 & A mismatch between the model and the parameters held by the optimizer causes a KeyError during initialization. \\
        DS-5489 & Freezing parameters prior to initializing DeepSpeed causes incomplete model checkpoints. \\
        DS-6714 & Using heterogeneous MoE architecture with pipeline parallelism causes inconsistent usage of communication primitives, leading to training stuck. \\
        DS-6772 & DeepSpeed initialization silently overwrites ``id'' attributes on models, causing wrong model-GPU placement. \\
        DS-6089 & The program is stuck on communication due to consistent ``capacity'' value across workers. \\
      \bottomrule
    \end{tabular}
    \caption{Six newly reported bugs that lead to silent errors, detected and diagnosed with the help of \sys. AC: Accelerate. DS: DeepSpeed. Numbers refer to GitHub issue IDs.}
    \label{tab:eval_new_bugs}
\end{table}

\subsection{New Silent Errors}
To further test \sys's effectiveness, we monitor recent open GitHub issues in
DeepSpeed and Transformers. We focus on issues that have silent symptoms, are
unresolved with unknown root causes, and reproducible. We create
reproduction scripts based on the reports to ensure the issues occur with the
latest library. We then apply the invariants \sys infers from the sample
pipelines to the issues.

During this exercise, \sys detects 6 new silent errors at an
early stage and aid in diagnosing their root causes, as summarized in
\autoref{tab:eval_new_bugs}. Three of these root causes have since been
confirmed and fixed.

\label{sec:eval:case-study}
\paragraph{Case Study: AC-2665} It causes the model to not learn effectively during
training. In the original issue report, the user adapted their original
pipeline into DDP, but the model stopped learning at all. The user did not know
the root cause, but found a setting that fixed the error, \inlinecode{use\_orig\_param\: true}.

We applied the invariants inferred from the GCN example to the user's pipeline
and inspected invariant violations. Below, we present three example true positives identified by the inferred invariants.
Inv1: \inlinecode{zero\_grad} should contain
changing of \inlinecode{grad} attributes from a non-zero tensor to a zero tensor or None.
Inv2: \inlinecode{step} should contain changing of \inlinecode{data} attributes of the model
parameters.  Inv3: \inlinecode{step} should contain a number of invocations of
mathematical operations on the model parameters, such as \inlinecode{\_foreach\_add}.
Inv2 and Inv3 indicate that the optimizer is not performing updates to the model, while Inv1
indicates that no gradient was computed. The three invariants combined
point out that the optimizer is likely not initialized with parameters that
are actually used during forward and backward passes.  We then checked the model
parameters and the optimizer's param\_groups and confirmed the hypothesis.
Upon investigation of the model and optimizer initialization process, we found
out that DDP automatically flattens the original model parameters and creates a
new model with the flattened parameters. However, the user initialized the
optimizer with the original model parameters, and thus the optimizer did not 
have the correct parameters to update. 
We reported the issue both to the user and to the transformers team,
and the issue has been confirmed. A pull request has been under review to fix the issue.

\subsection{False Positive}\label{subsec:fp}
To measure false positive, we collect \numofevalfpprogs diverse training
programs drawn from existing tutorials, all without known bugs. They span
different training scales (\emph{e.g.}, single-GPU vs.  multi-GPU), frameworks
(PyTorch, Transformers, Diffusers, \emph{etc.}), tasks (image classification,
language modeling, vision transformer pretraining, \emph{etc.}), and
configuration parameters (\emph{e.g.}, precision, batch size, dataset, and
model architecture).

\begin{figure}[t]
  \centering
  \includegraphics[width=\columnwidth]{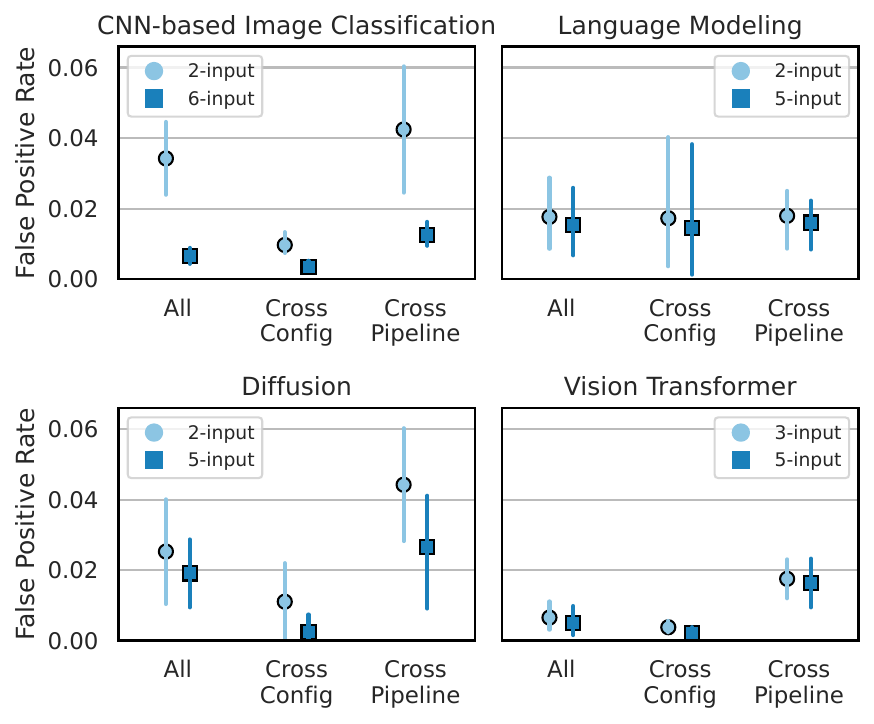}
  \caption{False positive rates across four program classes,
  broken down by cross-configuration and cross-pipeline settings.}
  \label{fig:false_positive_rates}
\end{figure}

To minimize confounding effects from applying invariants across unrelated
tasks, we group programs into four classes based on training task type, as
shown in \autoref{fig:false_positive_rates}.  Transferability across task
classes is evaluated separately in \autoref{subsec:transfer}.  For each class,
we split the programs into a training set (used to infer invariants) and a
validation set (used to assess false positives). Validation programs are
further categorized as either \emph{cross-configuration} (differing from the
training set only in configuration parameters) or \emph{cross-pipeline}
(different code with similar semantics).  This categorization
allows us to assess how well the inferred invariants generalize across both
minor and structural variations in training programs.

\autoref{fig:false_positive_rates} shows that \sys achieves consistently low
false positive rates.  In the primary evaluation setting, where invariants are
inferred from representative workloads using 5 or 6 input programs, false
positive rates remain below 2\% across all classes.  Even in constrained
settings with only 2 or 3 input programs, the rate stays below 5\%.

\subsection{Invariant Transferability}\label{subsec:transfer}

\sys infers transferable invariants, enabling those learned from a small set of 
input programs to generalize across different pipelines and library versions.
To evaluate this transferability, we apply a set of \emph{valid} invariants to
all \numofevalfpprogs collected pipelines. These invariants are inferred using
a 5/6-input setup across all classes and exclude any that triggered false
positives in \autoref{subsec:fp}. For each invariant, we count how many
pipelines it can be applied to without raising a false alarm.

\begin{figure}[t]
  \centering
  \includegraphics[width=\columnwidth]{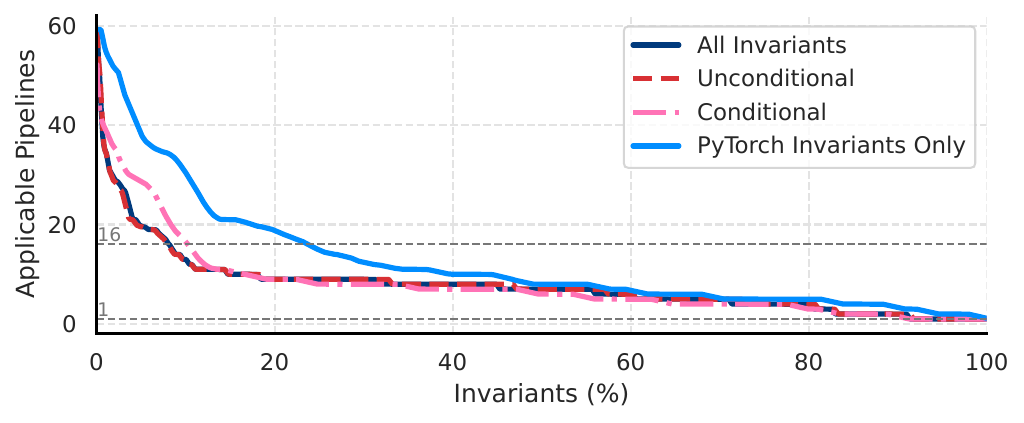}
  \caption{Invariant applicability across all collected pipelines. The y-axis
  shows how many pipelines each invariant applies to. ``All Invariants'' shows
  overall coverage; ``PyTorch Invariants Only'' isolates PyTorch-specific APIs.
  ``Conditional'' and ``Unconditional'' indicate whether preconditions are
  present.}
  \label{fig:transferability}
\end{figure}

As \autoref{fig:transferability} shows, many invariants exhibit broad
transferability. All invariants apply to at least one additional pipeline
beyond those used for inference.  Notably, over 8\% of invariants apply to more
than 16 pipelines---the average number of pipelines per model class,
demonstrating strong cross-class generalization despite semantic and structural
differences. We also observe that invariants with preconditions are generally
more transferable than unconditional ones, underscoring the importance of
precise precondition inference.

We expect the transferability to be even higher in practice, as developers
typically apply invariants only within the same framework or library context.
For example, invariants involving Transformers-specific APIs would not be
applied to a program that only uses PyTorch.  Since all pipelines in our
dataset use PyTorch, %
we isolate 8{,}172 invariants that capture only PyTorch-specific semantics.
These exhibit significantly higher transferability: 23\% of them apply to more
than 16 pipelines. This result suggests that framework-level behavior is a
strong source of reusable invariants. %

We also evaluate the false positive rate of applying invariants across model
classes to assess \sys's robustness in extreme transfer scenarios.  Using the
setup from \autoref{subsec:fp}, we infer invariants from one class and apply
them to all programs in the other classes, comparing the resulting false
positive rate to the in-class baseline.  %
Surprisingly, only the smallest-scale class, CNN, shows a higher false
positive rate in the cross-class setting (2.62\% vs. 0.65\%). In all other
cases, the rate is comparable or lower (\emph{e.g.}, 0.93\% vs. 1.54\% for
language modelling).  This is because many invariants become inapplicable due
to differing API usage and training context.

\subsection{False Negative}\label{subsec:fn}

To study the trade-off between false negative and the input programs used
for invariant inference, we evaluate three settings:
\emph{cross-configuration}, \emph{cross-pipeline}, and \emph{random}. In the
\emph{cross-configuration} setting, invariants are inferred from historical
runs of the same training pipeline executed under alternative configurations
where the silent error was not observed. In the \emph{cross-pipeline} setting,
invariants are inferred from semantically similar training pipelines that do
not exhibit the error.  In the \emph{random} setting, invariants are inferred
from general tutorial pipelines collected from relevant frameworks.

For each silent error detected in \autoref{sec:eval:issue-detection}, we
randomly sample $k$ input pipelines for each setting and evaluate whether the
error can be detected using the inferred invariants.  The average detection
rate for a given $k$ is computed by repeating this process 100 times and
averaging the results across all cases.

\begin{figure}[t]
  \centering
  \includegraphics[width=\columnwidth]{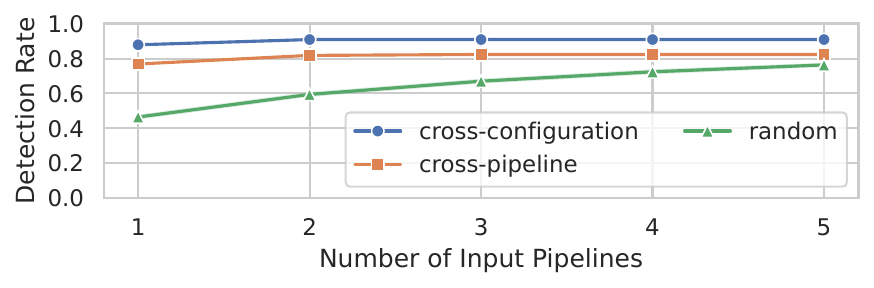}
  \caption{Detection rate as the number of input pipelines increases under cross-configuration, cross-pipeline, and random settings.}
  \label{fig:false_negative_curves}
\end{figure}

As \autoref{fig:false_negative_curves} shows, increasing the number of input
pipelines consistently improves detection rates. Both the cross-configuration
and cross-pipeline settings achieve high detection coverage (91\% and 82\%,
respectively) even with only two input programs.  The random setting starts
with lower detection coverage, but improves steadily as more inputs are added
(76\% with five inputs).  Further investigation of the undetected silent errors
reveals that the violated semantics often involve specialized features
that are underrepresented in the available example pipelines or not exercised
in cross-configuration or cross-pipeline inputs. For instance, detecting
\texttt{DeepSpeed-5794} requires invariants about DeepSpeed's MoE features;
however, only 1 out of 15 available DeepSpeed tutorial pipelines performs MoE
training, making it infeasible to detect \texttt{DeepSpeed-5794} with random
sampling.

\subsection{Invariant Sampling}
Sampling is necessary due to the large number of invariants inferred from the input programs.
For instance, six input programs can produce approximately 3,000 invariants covering various
APIs and variables in the PyTorch library. However, checking all these invariants is impractical,
and not all are equally important.

Many invariants pertain to well-tested APIs and variables that are unlikely to change, such as
\inlinecode{torch.nn.functional.relu}, \inlinecode{torch.cuda.is\_available}, and \inlinecode{torch.get\_default\_dtype}.
Additionally, some invariants are redundant, describing similar behaviors.
For example, invariants for \inlinecode{Adam.step} and \inlinecode{Adam.adam} overlap, as \inlinecode{Adam.step}
is a wrapper around \inlinecode{Adam.adam}.

To address this, we pruned the ~3,000 invariants down to 100,
focusing on the most critical APIs and variables in the PyTorch library,
prioritizing those less tested or more likely to change.

\subsection{Overhead}
\label{sec:eval-overhead}

\begin{figure}[t]
    \centering
    \includegraphics[width=\columnwidth]{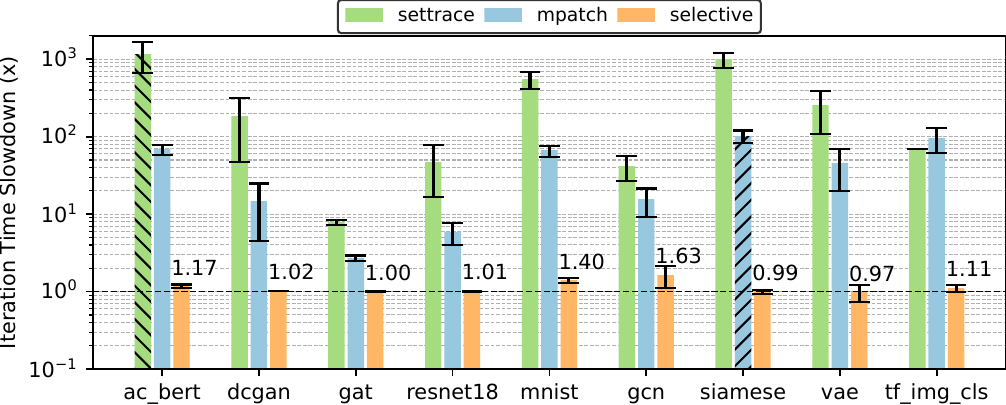}
    \caption{Overhead of different instrumentation techniques.}
    \label{fig:eval_instrument_overhead}
\end{figure}

We evaluate the runtime overhead of invariant checking during training.  \sys
uses \emph{selective instrumentation}, which only instruments the APIs and
variables relevant to the deployed invariants.  We conduct experiments on a
diverse set of training programs that vary in model size, task complexity, and
framework usage.  For each program, we deploy 100 randomly sampled invariants
and compare the per-iteration training time before and after instrumentation.
We compare our selective instrumentation against two baselines: (1) Python's
\emph{sys.settrace}, using a simple trace function that logs API calls and
arguments without variable tracking, and (2) \sys's \emph{full instrumentation}
mode, which instruments all API calls and variables regardless of invariant
relevance.

\autoref{fig:eval_instrument_overhead} shows the results.  \sys incurs low
overhead in selective mode, typically less than 2\%, and at most 1.6$\times$
slowdown across all workloads.  \inlinecode{GCN} and \inlinecode{MNIST} exhibit
higher relative overhead (1.6$\times$ and 1.4$\times$, respectively), as they
are toy workloads (\emph{e.g.}, training a 2-layer CNN) where per-iteration execution
time is minimal, and any instrumentation incurs a larger proportional cost.  In
contrast, for the more realistic workloads, overhead is significantly lower, as
a larger portion of the time is spent on GPU-bound computation.

The primary sources of overhead stem from trace data serialization into JSON,
conversion of objects into dictionary representations, and handling of tracked
objects. These operations occur at the Python level and introduce runtime costs,
especially in tight CPU-side loops.  While our current implementation is
synchronous and prioritizes correctness and modularity, there are clear
opportunities for reducing overhead further.  Potential engineering
optimizations include asynchronous or batched logging and minimizing redundant
instrumentation through static analysis.  We leave the exploration of these
optimizations to future work.

\begin{figure}[t]
    \centering
    \includegraphics[width=3.2in]{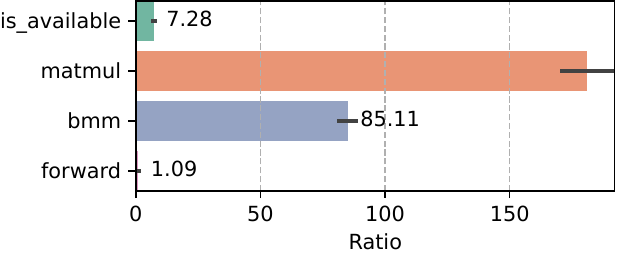}
    \caption{Microbenchmark of the overhead of the wrapper around the PyTorch APIs.
    APIs are listed from top to down by increasing order of its complexity.}
    \label{fig:eval_instrument_overhead_microbenchmark}
\end{figure}

We additionally conduct a microbenchmark to measure the overhead introduced by our wrapper around
PyTorch APIs. The overhead was quantified as the ratio of time spent in the wrapper
function to the time spent purely in the original function.
The results are presented in \autoref{fig:eval_instrument_overhead_microbenchmark}.

The relative overhead depends on two main factors:
\begin{itemize}[noitemsep, topsep=0pt, partopsep=0pt, leftmargin=*]
    \item \textbf{Function Complexity:} For APIs with similar interfaces, the wrapper's overhead remains nearly constant. Thus, the relative overhead is higher for lightweight functions, as their execution time is short, and lower for more computationally intensive functions.
    \item \textbf{Argument Complexity:} The wrapper's overhead grows linearly with the number and complexity of arguments. For example, dumping tensors is particularly costly due to the time required for serialization.
\end{itemize}

For lightweight APIs like \inlinecode{torch.matmul}, the wrapper's overhead is
significant because the function itself executes quickly, making the time spent
on dumping tensors disproportionately high.  In contrast, for simple functions
like \inlinecode{torch.cuda.is\_available}, which lack arguments, the overhead
is negligible.  Similarly, for higher-level APIs such as
\inlinecode{module.forward}, which invoke many lower-level APIs and have long
execution times, the wrapper's overhead is amortized over the duration of the
original function.  These observations suggest avenues for future work, such as
\emph{Inference Prioritization} using static analysis or GitHub commit
histories to understand which APIs are more likely to change and should be
prioritized for invariant checking, and \emph{Overhead Optimization} by
balancing the granularity of checking with the minimum information required to
infer effective invariants.

\subsection{Inference Efficiency}
\label{subsec:inference-efficiency}

To quantify the invariant inference efficiency, we reuse the traces
from the false positive experiments (\autoref{subsec:fp}) and measure
the inference time across traces of varying size and structural complexity.
For consistency, we normalize trace size by treating the trace from a ResNet-18
pretraining run with 20 training iterations and 10 testing iterations as a
standard program trace, corresponding to 66.2 MB or 93,686 records.

\begin{figure}[t]
  \centering
  \includegraphics[width=\columnwidth]{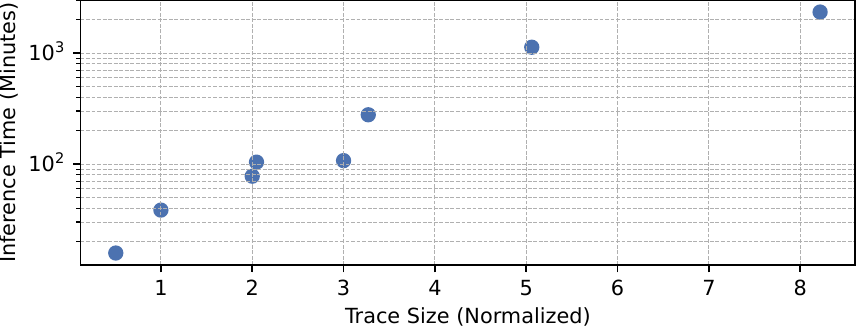}
  \caption{Inference time vs. trace size (normalized to ResNet-18 pretraining trace size).}
  \label{fig:inference_cost}
\end{figure}

\autoref{fig:inference_cost} shows that inference time grows roughly quadratically with trace size in this setting. 
While the inference algorithm itself is linear in both trace size and the
number of hypotheses, larger traces typically expose more semantic behaviors,
resulting in a larger hypothesis set.  In the worst case, \sys completes
inference in 38 hours when processing traces from 8.2 standard programs. Since
inference runs offline and our current implementation is single-threaded, the
performance remains acceptable for practical usage.

\subsection{Examining Invariant Violations}

Despite a low false positive rate, false alarms do occur, and developers must
examine violation reports to extract actionable diagnostic information.  We
find that violations can often be inspected structurally rather than treated in
isolation.  They tend to cluster around specific APIs or components, which
makes the review process more manageable.  In practice, true positives are
frequently supported by multiple related violations that reinforce the
underlying issue, while false positives follow recognizable patterns and are
easy to dismiss.

For example, in the AC-2665 case study (\autoref{sec:eval:case-study}), using invariants inferred from PyTorch's GCN pipeline alone, 100 violations were reported. 
Among these, 52 were true positives. 
Of those, 33 indicated that \inlinecode{torch.optim.adamw.adamw} was never invoked, highlighting a missing optimizer initialization.
18 showed that \inlinecode{optimizer.step} performed no mathematical operations, suggesting it was not linked to model parameters. 
The remaining 48 violations were quickly dismissed as irrelevant: 7 involved GCN-specific constants (e.g., \inlinecode{dropout\_rate == 0.5}), 
and 26 flagged missing ReLU invocations, which do not apply to the T5-based model used in AC-2665. 
The rest followed similar, non-impactful patterns.

\section{Limitations}

\tool has a few limitations. 
First, its instrumentation interferes with JIT compilation tools like 
\inlinecode{torch.compile}, preventing the analysis of optimized code paths.
Second, it is restricted to Python code, limiting its ability to analyze 
components with significant logic implemented in lower-level languages, 
such as the Flash Attention algorithm~\cite{dao2022flashattentionfastmemoryefficientexact}.
Finally, representing tensors in hash form prevents fine-grained numerical 
analysis, limiting its applicability to detecting instabilities caused by 
inappropriate hyperparameters. However, this can be complemented by existing 
research on hyperparameter tuning~\cite{yu2020hyperparameteroptimizationreviewalgorithms} 
and numerical defect detection~\cite{li2023reliability}.

\section{Related Work}
\label{sec:related}

\paragraph{Testing Pipeline Code, Libraries, and Compilers}
 CRADLE~\cite{pham_cradle_2019}, AUDEE~\cite{guo_audee_2020}, LEMON~\cite{10.1145/3368089.3409761}, 
and NNSmith~\cite{NNSmith2023ASPLOS} employ differential testing to detect and diagnose failures across frameworks and compilers.
PyTea~\cite{PyTea2022ICSE} detects tensor shape mismatches using pre-specified API constraints, 
while NeuRI~\cite{liu2023neuridiversifyingdnngeneration} enhances this by automating constraint inference. 
RANUM~\cite{li2023reliability} targets numerical defects in deep neural networks. 
\tool goes beyond framework inconsistencies or isolated error categories by inferring runtime invariants tailored to the training process. 
It automates error detection and debugging with automated invariant inference from example pipelines, 
addressing a broader range of silent failures across diverse training pipelines.

\paragraph{Monitoring Frameworks for Training Dynamics}  
Tools like TensorBoard~\cite{TensorFlow2016OSDI} and Weights \&
Biases~\cite{wandb} log high-level metrics such as loss and accuracy, enabling
developers to visualize and compare experiments easily.  However, these tools
require manual and active monitoring, as seen during the training of BloombergGPT
\cite{wu2023bloomberggptlargelanguagemodel}, where loss plateaued for seven
days before the developer noticed it. These metrics are often noisy 
and can lead to many false alarms when used for detecting silent errors. They also do
not help with diagnosis. 

In contrast, \tool provides an automated solution that captures the precise
semantics of training as training invariants and proactively
checks them, enabling reduced manual effort, early and accurate detection of
silent training errors, and providing diagnosis hints for root-cause analysis.

\paragraph{Testing DL Models}  
Extensive work exists to test trained DL models for robustness and fairness,
such as DeepXplore~\cite{Pei_2017} and DeepTest~\cite{tian2018deeptest}. 
They are orthogonal to \sys given their focus on testing and uncovering errors
in the final model weights rather than validating the training process.

\paragraph{Fault-Tolerance in DL Systems}  
Fault-tolerance mechanisms have been proposed, such as elastic resource scaling, task reallocation, 
pipeline parallelism, and efficient checkpointing~\cite{Oobleck2023SOSP, Bamboo2023NSDI, Varuna2022EuroSys, lian2024universalcheckpointingefficientflexible}.
While these approaches improve fault tolerance, they do not address silent errors that arise from misconfigurations, bugs, or subtle correctness violations.

\paragraph{Invariant Mining}
Much work has explored mining likely program invariants for traditional
single-component software with tools such as Daikon~\cite{Daikon1999ICSE} and
DIDUCE~\cite{DIDUCE2002ICSE}. These invariants focus on low-level program
variable relations at certain points of the program, such as \texttt{idx < len}
for two local variables in a loop. Recent work such as I4~\cite{I42019SOSP},
DistAI~\cite{DistAI2021OSDI}, and DuoAI~\cite{DuoAI2022OSDI} infer inductive
invariants, which are used in the verification of distributed protocols.
Oathkeeper~\cite{OKLib2022OSDI} infers event rules to detect silent failures in
distributed systems.

Inferring rules is a general approach for bug detection.  Engler \emph{et al.}~\cite{Engler2001Bugs} 
notably propose inferring rules about programmer beliefs and show its
effectiveness in large systems code.

\tool targets a new domain of deep learning training systems and addresses
various challenges unique to this domain. To the best of our knowledge, it is
the first work for systematic invariant checking in DL training pipelines to
detect silent training errors. The training invariants \tool infers capture
high-level semantics tailored to DL training. \tool also deduces
precise preconditions for these invariants. Moreover, prior work mainly infers
invariants from a single system, and these invariants only apply to that system.
In contrast, \tool invariants from diverse and seemingly unrelated training
pipelines. Its inferred invariants are transferable to different pipelines.

\section{Conclusion}
\label{sec:conclusion}
Silent errors are detrimental to DL training and yet notoriously difficult to
address. This paper presents a study on such errors to understand their
characteristics. Informed by the study, we propose a principle approach that
uses precise training invariants to detect and diagnose silent training errors.
We present \sys, an end-to-end framework that automates the process of
inferring training invariants and proactively checking them. \sys shows
effective and quick detection capability for real-world silent training errors
with diverse root causes. It also uncovers previously unknown
bugs leading to silent errors in popular training libraries. Its tailored
approach and the transferability of its inferred invariants make it readily
applicable to existing DL training pipelines. 

\section*{Acknowledgments}
\label{sec:ack}
We thank our shepherd and the anonymous reviewers for their constructive
feedback and guidance, which greatly improved the quality of the paper.  We are
grateful to Stas Bekman for sharing detailed insights into the Bloom-176B
training issue. We thank Yijun Wang for his contribution to the empirical
study. We appreciate the feedback from the members of OrderLab and Mosharaf Chowdhury.
We thank the Chameleon Cloud for providing experiment machine resources. This work
was supported in part by NSF grants CNS-2317698, CNS-2317751, and CCF-2318937.

{
\bibliographystyle{plain}

}

\end{document}